\documentclass{article}

\usepackage[final]{corl_2020} %

\usepackage{algorithm}
\usepackage{algorithmic}
\usepackage{float}
\usepackage{bbm}
\usepackage{graphicx}
\usepackage{subcaption}
\usepackage{xcolor}
\usepackage{hyperref}       %
\usepackage{url}            %
\usepackage{booktabs}       %
\usepackage{amsfonts}       %
\usepackage{amsmath,amssymb}
\usepackage{nicefrac}       %
\usepackage{enumitem}
\usepackage{wrapfig}

\newcommand{\Skip}[1]{}

\newcommand{\ie}{i.e.,\ }
\newcommand{\eg}{e.g.,\ }

\newcommand{\mysecref}[1]{Section~\ref{#1}}
\newcommand{\myfigref}[1]{Figure~\ref{#1}}

\newcommand{\myeqref}[1]{Equation~\ref{#1}}
\newcommand{\myalgref}[1]{Algorithm~\ref{#1}}

\renewcommand{\algorithmiccomment}[1]{\hfill\footnotesize{\textcolor{gray}{$\triangleright$ #1}}}

\title{Motion Planner Augmented Reinforcement Learning \\ for Robot Manipulation in Obstructed Environments}

\author{
  Jun Yamada$^1$\thanks{
  Equal contribution. Correspondence to:  \texttt{\href{mailto:jy\_597@usc.edu}{jy\_597@usc.edu}} and  \texttt{\href{mailto:lee504@usc.edu}{lee504@usc.edu}}} , 
  Youngwoon Lee$^1$\footnotemark[1] , 
  Gautam Salhotra$^2$, 
  Karl Pertsch$^1$, \\
  \textbf{Max Pflueger$^2$, 
  Gaurav S. Sukhatme$^2$, 
  Joseph J. Lim$^1$, 
  Peter Englert$^2$}\\
  $^1$ Cognitive Learning for Vision and Robotics Lab\\
  $^2$ Robotic Embedded Systems Laboratory\\ 
  Department of Computer Science\\
  University of Southern California
}

\begin{document}
\maketitle

\begin{abstract}
Deep reinforcement learning (RL) agents are able to learn contact-rich manipulation tasks by maximizing a reward signal, but require large amounts of experience, especially in environments with many obstacles that complicate exploration. In contrast, motion planners use explicit models of the agent and environment to plan collision-free paths to faraway goals, but suffer from inaccurate models in tasks that require contacts with the environment. To combine the benefits of both approaches, we propose motion planner augmented RL (MoPA-RL) which augments the action space of an RL agent with the long-horizon planning capabilities of motion planners. Based on the magnitude of the action, our approach smoothly transitions between directly executing the action and invoking a motion planner. We evaluate our approach on various simulated manipulation tasks and compare it to alternative action spaces in terms of learning efficiency and safety. The experiments demonstrate that MoPA-RL increases learning efficiency, leads to a faster exploration, and results in safer policies that avoid collisions with the environment. Videos and code are available at \url{https://clvrai.com/mopa-rl}.
\end{abstract}


\section{Introduction}
\begin{wrapfigure}{r}{0.375\textwidth}
    \centering
    \vspace{-1.5em}
    \includegraphics[width=1.0\linewidth]{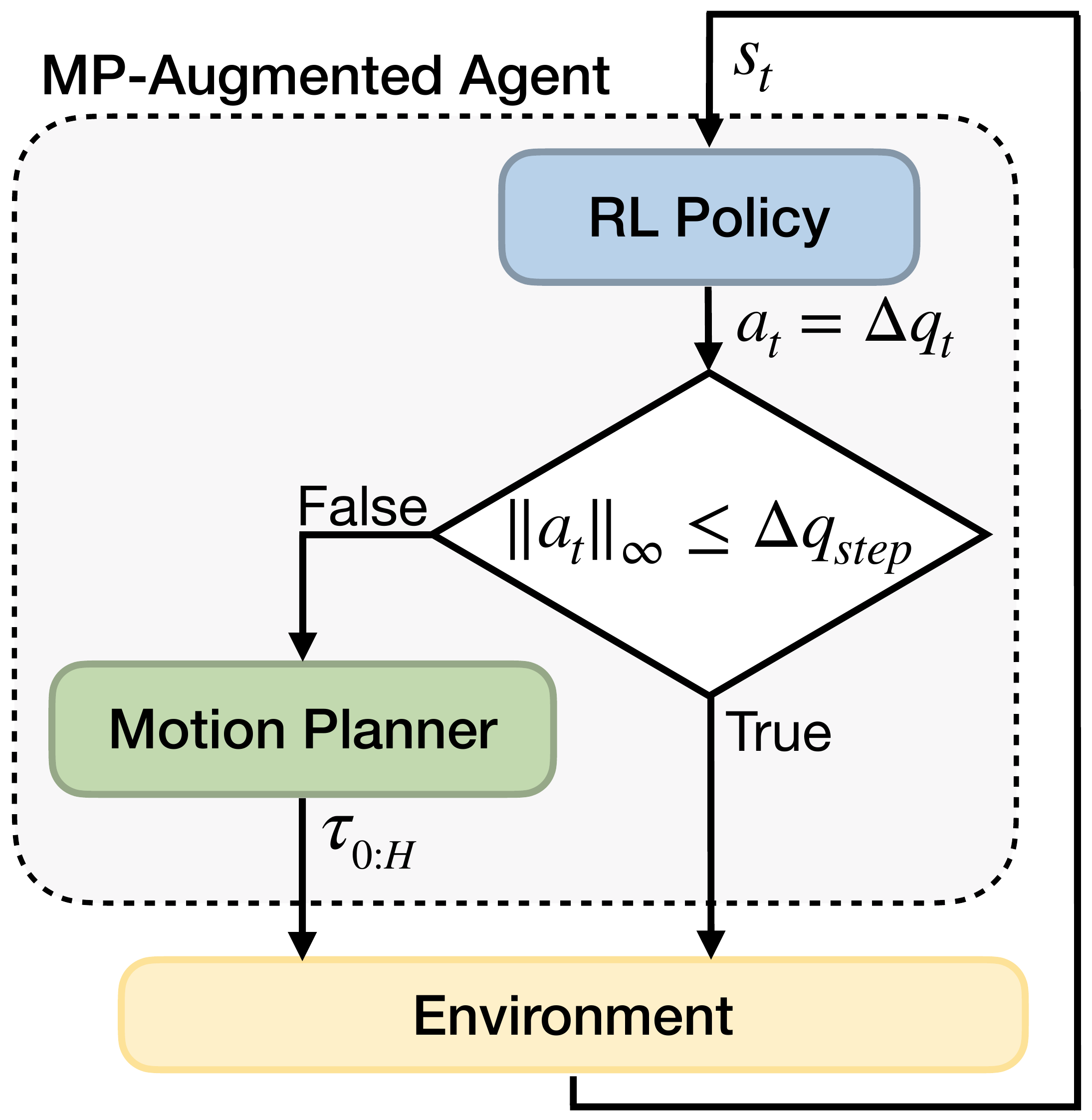}
    \caption{
        Our framework extends an RL policy with a motion planner. If the predicted action by the policy is above a threshold $\Delta q_\text{step}$, the motion planner is called; otherwise, it is directly executed.
    }
    \vspace{-1.5em}
    \label{fig:model}
\end{wrapfigure}

In recent years, deep reinforcement learning (RL) has shown promising results in continuous control problems~\citep{lillicrap2016ddpg, schulman2017ppo, haarnoja2018sac, lee2019composing, lee2020coordination}. Driven by rewards, robotic agents can learn tasks such as grasping~\citep{levine2016learning, kalashnikov2018qtopt} and peg insertion~\citep{fan2018surreal}. 
However, prior works mostly operated in controlled and uncluttered environments, whereas in real-world environments, it is common to have many objects unrelated to the task, which makes exploration challenging.
This problem is exacerbated in situations where feedback is scarce and considerable exploration is required before a learning signal is received.

Motion planning (MP) is an alternative for performing robot tasks in complex environments, and has been widely studied in the robotics literature~\citep{amato1996PRM, lavalle1998RRT, karaman2011RRTStar, elbanhawi2014samplingMPReview}.
MP methods, such as RRT~\cite{lavalle1998RRT} and PRM~\cite{amato1996PRM}, can find a collision-free path between two robot states in an obstructed environment using explicit models of the robot and the environment.
However, MP struggles on tasks that involve rich interactions with objects or other agents since it is challenging to obtain accurate contact models. Furthermore, MP methods cannot generate plans for complex manipulation tasks (\eg object pushing) that cannot be simply specified by a single goal state. 

In this work, we propose motion planner augmented RL (MoPA-RL) which combines the strengths of both MP and RL by augmenting the action space of an RL agent with the capabilities of a motion planner.
Concretely, our approach trains a model-free RL agent that controls a robot by predicting state changes in joint space, where an action with a large joint displacement is realized using a motion planner while a small action is directly executed. 
By predicting a small action, the agent can perform sophisticated and contact-rich manipulation. On the other hand, a large action allows the agent to efficiently explore an obstructed environment with collision-free paths computed by MP.

Our approach has three benefits: (1) MoPA-RL can add motion planning capabilities to \emph{any} RL agent with joint space control as it does not require changes to the agent's architecture or training algorithm; (2) MoPA-RL allows an agent to freely switch between MP and direct action execution by controlling the scale of action; and (3) the agent naturally learns trajectories that avoid collisions by leveraging motion planning, allowing for safe execution even in obstructed environments.

The main contribution of this paper is a framework augmenting an RL agent with a motion planner, which enables effective and safe exploration in obstructed environments. In addition, we propose three challenging robotic manipulation tasks with the additional challenges of collision-avoidance and exploration in obstructed environments.
We show that the proposed MoPA-RL learns to solve manipulation tasks in these obstructed environments while model-free RL agents suffer from local optima and difficult exploration.

\begin{figure*}
    \centering
    \includegraphics[width=\linewidth]{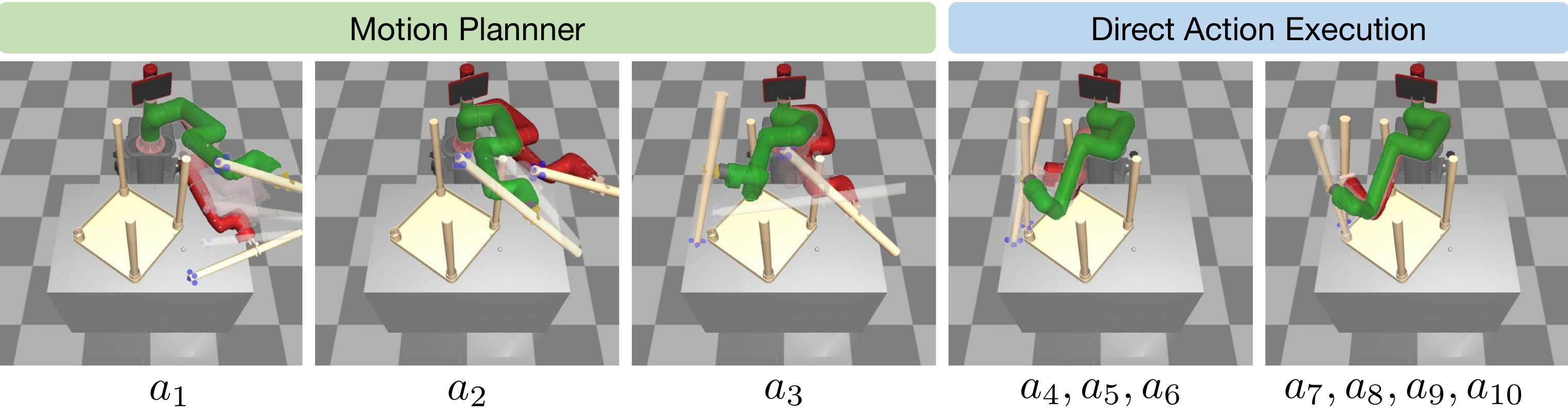}
    \caption{
        Learning manipulation skills in an obstructed environment is challenging due to frequent collisions and narrow passages amongst obstacles. 
        For example, when a robot moves the table leg to assemble, it is likely to collide with other legs and get stuck between legs.
        Moreover, once the table leg is moved, the robot is tasked to insert the leg into the hole, which requires contact-rich interactions.
        To learn both collision-avoidance and contact-rich skills, our method (MoPA-RL) combines motion planning and model-free RL. In the images, the green robot visualizes the target state provided by the policy.
        Initially, motion planning can be used to navigate to target states $a_1$, $a_2$, and $a_3$ while avoiding collision.
        Once the arm passes over other legs, a sequence of primitive actions $a_4 - a_{10}$ are directly executed to assemble the leg and tabletop.
    }
    \vspace{-8pt}
    \label{fig:execution}
\end{figure*}

\section{Related Work}
Controlling robots to solve complex manipulation tasks using deep RL~\citep{gu2016deep, levine2015endtoend, levine2013GPS, haarnoja2018sac} has been an active research area. Specifically, model-free RL~\citep{gu2016deep, levine2015endtoend, haarnoja2018sac} has been well studied for robotic manipulation tasks, such as picking and placing objects~\citep{zhu2018reinforcement}, in-hand dexterous manipulation~\citep{andrychowicz2018learning, rajeswaran2018learning}, and peg insertion~\citep{chebotar2019closing}. To tackle long-horizon tasks, hierarchical RL (HRL) approaches have extended RL algorithms with temporal abstractions, such as options~\citep{sutton1999between, bacon2017option}, modular networks~\citep{andreas2017modular, lee2019composing, lee2020coordination}, and goal-conditioned low-level policies~\citep{nachum2018data}. However, these approaches are typically tested on controlled, uncluttered environments and often require a large number of samples. 

On the other hand, motion planning~\citep{amato1996PRM, kavraki1994randomized, overmars1992random, lavalle1998rapidly, lavalle2000rapidly, karaman2011sampling}, a cornerstone of robot motion generation, can generate a collision-free path in cluttered environments using explicit models of the robot and environment. Probabilistic roadmaps (PRMs)~\citep{amato1996PRM, kavraki1994randomized, overmars1992random} and rapidly-exploring random trees (RRTs)~\citep{lavalle1998rapidly, lavalle2000rapidly, karaman2011sampling} are two common sampling-based motion planning techniques. These planning approaches can effectively compute a collision-free path in static environments; however, these methods have difficulties handling dynamic environments and contact-rich interactions, which frequently occur in object manipulation tasks.

There are several works that combine motion planning and reinforcement learning to get the advantages of both approaches.
A typical approach is to decompose the problem into the parts that can and cannot be solved with MP. Then, RL is used to learn the part that the planner cannot handle~\cite{englert2018ijrr, vuga2015enhanced, singh1994robust, chiang2019learning, xiali2020relmogen}.  
However, this separation often relies on
task-specific heuristics that are only valid for a limited task range. 
MP can be incorporated with RL in the form of a modular framework with a task-specific module switching rule~\citep{lee2020guided} and HRL with pre-specified goals for the motion planner~\citep{angelov2020composing}. 
\citet{xiali2020relmogen} uses a motion planner as a low-level controller and learn an RL policy in a high-level action (subgoal) space, which limits the capability of learning contact-rich skills.
Instead, we propose to \emph{learn} how to balance the use of the motion planner and primitive action (direct action execution) using model-free RL with minimal task-specific knowledge.

\begin{figure}
  \centering
  \begin{subfigure}{0.24\textwidth}
      \includegraphics[width=\textwidth]{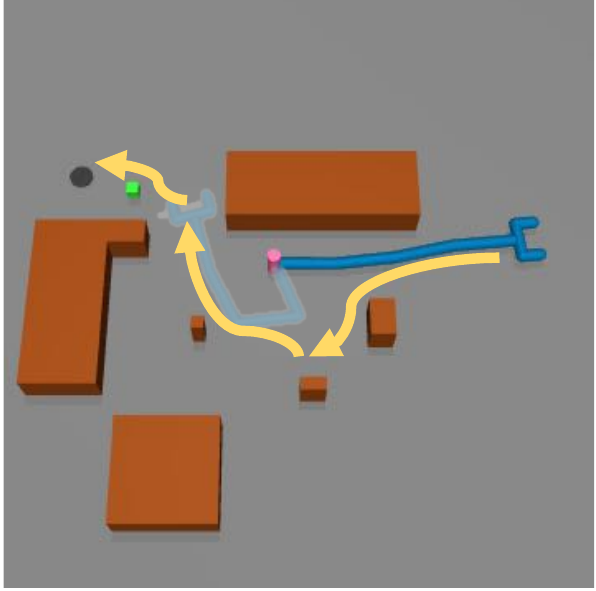}
      \caption{2D Push}
  \end{subfigure}
  \hfill
  \begin{subfigure}{0.24\textwidth}
      \includegraphics[width=\textwidth]{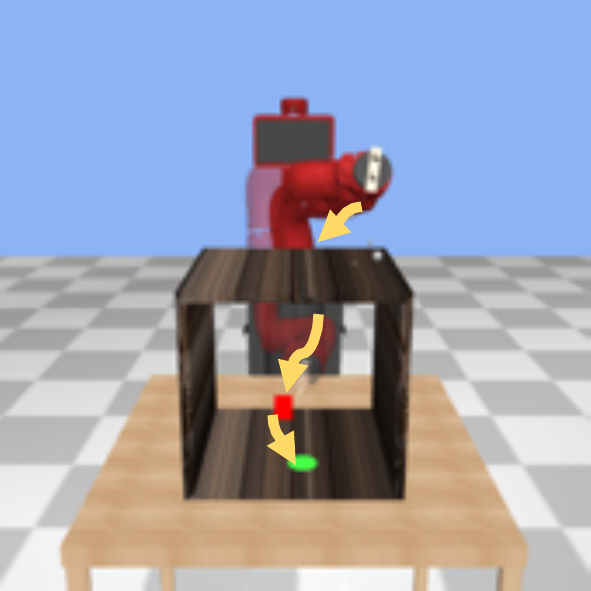}
      \caption{Sawyer Push}
  \end{subfigure}
  \hfill
  \begin{subfigure}{0.24\textwidth}
      \includegraphics[width=\textwidth]{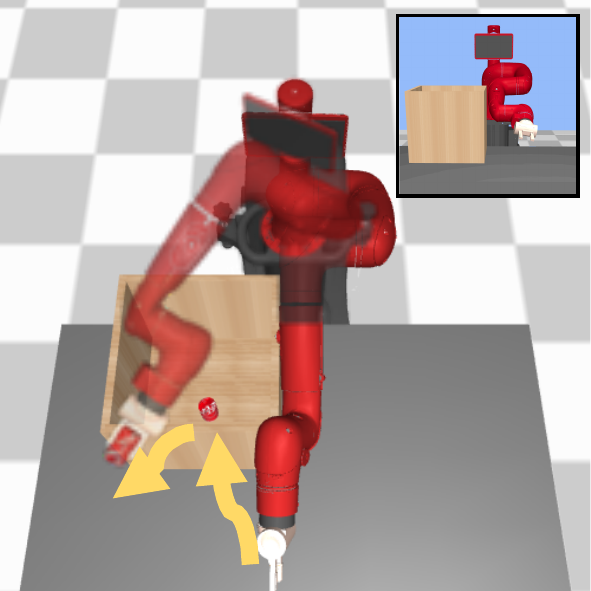}
      \caption{Sawyer Lift}
      \label{fig:env_sawyer_lift}
  \end{subfigure} 
  \hfill
  \begin{subfigure}{0.24\textwidth}
      \includegraphics[width=\textwidth]{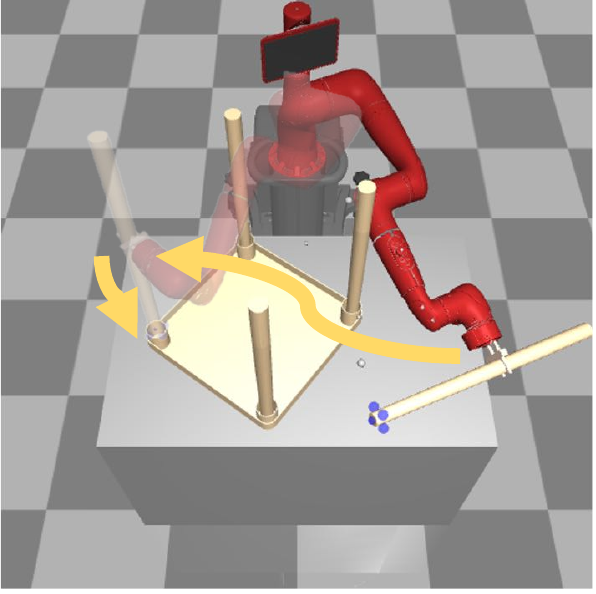}
      \caption{Sawyer Assembly}
      \label{fig:env_sawyer_assembly}
  \end{subfigure}
\caption{
Manipulation tasks in obstructed environments. (a)~\textit{2D Push}: The 2D reacher agent has to push the green box to the goal (black circle). (b)~\textit{Sawyer Push}: Sawyer arm should push the red box toward the goal (green circle). (c)~\textit{Sawyer Lift}: Sawyer arm takes out the can from the long box. (d)~\textit{Sawyer Assembly}: Sawyer arm moves and inserts the table leg into the hole in the table top.
}
\label{fig:envs}
\end{figure}
\section{Motion Planner Augmented Reinforcement Learning}

In this paper, we address the problem of solving manipulation tasks in the presence of obstacles. Exploration by deep reinforcement learning (RL) approaches for robotic control mostly relies on small perturbations in the action space. However, RL agents struggle to find a path to the goal in obstructed environments due to collisions and narrow passages. Therefore, we propose to harness motion planning (MP) techniques for RL agents by augmenting the action space with a motion planner. In \mysecref{sec:mopa-rl}, we describe our framework, MoPA-RL, in detail. Afterwards, we elaborate on the motion planner implementation and RL agent training.

\subsection{Preliminaries}
\label{sec:prelim}
We formulate the problem as a Markov decision process (MDP) defined by a tuple $(\mathcal{S}, \mathcal{A}, P, R, \rho_0, \gamma)$ consisting of states $s\in\mathcal{S}$, actions $a\in\mathcal{A}$, transition function $P(s'\in\mathcal{S}|s,a)$, reward $R(s,a)$, initial state distribution $\rho_0$, and discount factor $\gamma \in [0, 1]$. 
The agent's action distribution at time step $t$ is represented by a policy $\pi_{\phi}(a_t|s_t)$ with state $s_t \in \mathcal{S}$ and action $a_t \in \mathcal{A}$, where $\phi$ denotes the parameters of the policy. 
Once the agent executes the action $a_t$, it receives a reward $r_t=R(s_t,a_t)$.
The performance of the agent is evaluated using the discounted sum of rewards $\sum_{t=0}^{T-1} \gamma^t R(s_t, a_t)$, where $T$ denotes the episode horizon.

In continuous control, the action space can be defined as the joint displacement $a_t = \Delta q_t$, where $q_t$ represents robot joint angles. To prevent collision and reduce control errors, the action space is constrained to be small, $\mathcal{A} = [-\Delta q_\text{step}, \Delta q_\text{step}]^d$, where $\Delta q_\text{step}$ represents the maximum joint displacement for a direct action execution~\citep{gu2016deep} and $d$ denotes the dimensionality of the action space.

On the other hand, a kinematic motion planner $\textrm{MP}(q_t,g_t)$ computes a collision-free path $\tau_{0:H}=(q_t, q_{t+1}, \dots, q_{t+H}=g_t)$ from a start joint state $q_t$ to a goal joint state $g_t$, where $H$ is the number of states in the path. 
The sequence of actions $a_{t:t+H-1}$ that realize the path $\tau_{0:H}$ can be obtained by computing the displacement between consecutive joint states, $\Delta \tau_{0:H}=(\Delta q_t, \dots, \Delta q_{t+H-1})$.

\begin{algorithm}[t]
    \caption{Motion Planner Augmented RL (MoPA-RL)}
    \label{alg:training}
    \begin{algorithmic}[1]
        \renewcommand{\COMMENT}[1]{{\algorithmiccomment{#1}}}
        \REQUIRE Motion planner $\textrm{MP}$, augmented MDP $\tilde{\mathcal{M}}(\mathcal{S}, \tilde{\mathcal{A}}, \tilde{P}, \tilde{R}, \rho_0, \gamma)$, action limits $\Delta q_\text{step}, \Delta q_\text{MP}$
        number of reused trajectories $M$
        
        \STATE Initialize policy $\pi_\phi$ and replay buffer $\mathcal{D}$
        \FOR{$i=1,2,\dots$}
            \STATE Initialize episode $s_0\sim \rho_0, \tilde{t} \leftarrow 0, t \leftarrow 0$
            \WHILE{episode not terminated}
                \STATE $\tilde{a}_{\tilde{t}} \sim \pi_{\phi}(\tilde{a}_{\tilde{t}}|s_{t})$
                \IF{$||\tilde{a}_{\tilde{t}}||_{\infty} > \Delta q_\text{step}$}     
                    \STATE $H_{\tilde{t}}, \tau_{0:H_{\tilde{t}}} \leftarrow \textrm{MP}(q_t, q_t + \tilde{a}_{\tilde{t}})$ \COMMENT{Motion planner execution}
                    \STATE $s_{t+H_{\tilde{t}}}, \tilde{r}_{\tilde{t}} \leftarrow \tilde{P}(s_t, \Delta \tau_{0:H_{\tilde{t}}}), \tilde{R}(s_t, \Delta  \tau_{0:H_{\tilde{t}}})$
                    \FOR{$j=1,\dots,M$}
                        \Skip{
                        \STATE $n\sim\mathcal{U}\{0, H_{\tilde{t}}-1\}, m\sim\mathcal{U}\{n+1, H_{\tilde{t}}\}$
                        \STATE $\tilde{a}, \tilde{r}\leftarrow \tau_m - \tau_n, \tilde{R}(\tau_{n}, \Delta\tau_{n:m})$
                        \STATE $\mathcal{D} \leftarrow \mathcal{D} \cup \{(\tau_n, \tilde{a}, \tilde{r}, \tau_m, m-n)\}$ \COMMENT{Reuse intermediate trajectories}
                        }
                        \STATE Sample intermediate transitions $\tau_{n:m}$ from $\tau_{0:H_{\tilde{t}}}$
                        \STATE $\tilde{a}, \tilde{r} \leftarrow \Delta \tau_{n:m}, \tilde{R}(s_{t+n}, \Delta \tau_{n:m})$
                        \STATE $\mathcal{D} \leftarrow \mathcal{D} \cup \{(s_{t+n}, \tilde{a}, \tilde{r}, s_{t+m}, m-n)\}$  \COMMENT{Reuse motion plan trajectories}        
                    \ENDFOR
                \ELSE            
                    \STATE $H_{\tilde{t}} \leftarrow 1$
                    \STATE $s_{t+H_{\tilde{t}}}, \tilde{r}_{\tilde{t}} \leftarrow \tilde{P}(s_t, \tilde{a}_{\tilde{t}}), \tilde{R}(s_t, \tilde{a}_{\tilde{t}})$ \COMMENT{Direct action execution}
                \ENDIF
                \STATE $\mathcal{D} \leftarrow \mathcal{D} \cup \{(s_t, \tilde{a}_{{\tilde{t}}}, \tilde{r}_{\tilde{t}}, s_{t+H_{\tilde{t}}}, H_{\tilde{t}})\}$
                \STATE $t \leftarrow t+H_{\tilde{t}}, \tilde{t} \leftarrow \tilde{t} + 1$
                \STATE Update $\pi_{\phi}$ using model-free RL
            \ENDWHILE
        \ENDFOR
    \end{algorithmic}
\end{algorithm}

\subsection{Motion Planner Augmented Reinforcement Learning}
\label{sec:mopa-rl}

To efficiently learn a manipulation task in an obstructed environment, we propose motion planner augmented RL (MoPA-RL). Our method harnesses a motion planner for controlling a robot toward a faraway goal without colliding with obstacles, while directly executing small actions for sophisticated manipulation.
By utilizing MP, the robot can effectively explore the environment avoiding obstacles and passing through narrow passages. For contact-rich tasks where MP often fails due to an inaccurate contact model, actions can be directly executed instead of calling a planner.

As illustrated in \myfigref{fig:model}, our framework consists of two components: an RL policy $\pi_\phi(a|s)$ and a motion planner $\textrm{MP}(q,g)$.
In our framework, the motion planner is integrated into the RL policy by enlarging its action space.
The agent directly executes an action if it is in the original action space. If an action is sampled from outside of the original action space, which requires a large movement of the agent, the motion planner is called and computes a path to realize the large joint displacement.

To integrate the motion planner with an MDP $\mathcal{M}$, we first define an augmented MDP $\tilde{\mathcal{M}}(\mathcal{S}, \tilde{\mathcal{A}}, \tilde{P}, \tilde{R}, \rho_0, \gamma)$, where $\tilde{\mathcal{A}} = [-\Delta q_\text{MP}, \Delta q_\text{MP}]^d$ is an enlarged action space with $\Delta q_\text{MP} > \Delta q_\text{step}$, $\tilde{P}(s'|s,\tilde{a})$ denotes the augmented transition function, and $\tilde{R}(s,\tilde{a})$ is the augmented reward function. 
Since one motion planner call can execute a sequence of actions in the original MDP $\mathcal{M}$, the augmented MDP $\tilde{\mathcal{M}}$ can be considered as a semi-MDP~\citep{sutton1999between}, where an option $\tilde{a}$ executes an action sequence $\Delta \tau_{0:H}$ computed by the motion planner. 
For simplicity of notation, we use $\tilde{a}$ and $\Delta \tau_{0:H}$ interchangeably.
The augmented transition function $\tilde{P}(s'|s,\tilde{a})=\tilde{P}(s'|s,\Delta \tau_{0:H})$ is the state distribution after taking a sequence of actions and the augmented reward function $\tilde{R}(s,\tilde{a})=\tilde{R}(s,\Delta \tau_{0:H})$ is the discounted sum of rewards along the path.

On the augmented MDP $\tilde{\mathcal{M}}$, the policy $\pi_\phi(\tilde{a}| s)$ chooses an action $\tilde{a}$, which represents a change in the joint state $\Delta q$. The decision whether to call the motion planner or directly execute the predicted action is based on its maximum magnitude $||\tilde{a}||_{\infty}$, \ie the maximum size of the predicted displacement, as illustrated in \myfigref{fig:model}. If the joint displacement is larger than an action threshold $\Delta q_\text{step}$ for any joint (\ie $||\tilde{a}||_{\infty} > \Delta q_\text{step}$), which is likely to lead to collisions, the motion planner is used to compute a collision-free path $\tau_{0:H}$ towards the goal $g = q + \tilde{a}$. 
To follow the path, the agent executes the action sequence $\Delta \tau_{0:H}$ over $H$ time steps. 
Otherwise, \ie $||\tilde{a}||_{\infty} \leq \Delta q_\text{step}$, the action is directly executed using a feedback controller for a single time step, as is common practice in model-free RL. 
This procedure is repeated until the episode is terminated. Then, the policy $\pi_\phi$ is trained to maximize the expected returns $\mathbb{E}_{\pi_\phi}\left[\sum_{\tilde{t}=0}^{\tilde{T}} \gamma^t \tilde{R}(s_{\tilde{t}}, \tilde{a}_{\tilde{t}})\right]$, where $\tilde{T}$ is the episode horizon on $\tilde{\mathcal{M}}$ and $t=\sum_{i=0}^{\tilde{t}-1} H_i$ is the number of primitive actions executed before time step $\tilde{t}$.
The complete RL training with the motion-planner augmented agent is described in \myalgref{alg:training}.

The proposed method has three advantages. First, our method gives the policy the freedom to choose whether to call the motion planner or directly execute actions by predicting large or small actions, respectively. Second, the agent naturally produces trajectories that avoid collisions by leveraging MP, allowing for safe policy execution even in obstructed environments. The third advantage is that MoPA-RL can add motion planning capabilities to \emph{any} RL algorithm with joint space control as it does not require changes to the agent's architecture or training procedure.

\subsection{Action Space Rescaling}
\label{sec:rescaling}

\begin{wrapfigure}{r}{0.4\textwidth}
    \centering
    \vspace{-10pt}
    \includegraphics[width=0.9\linewidth]{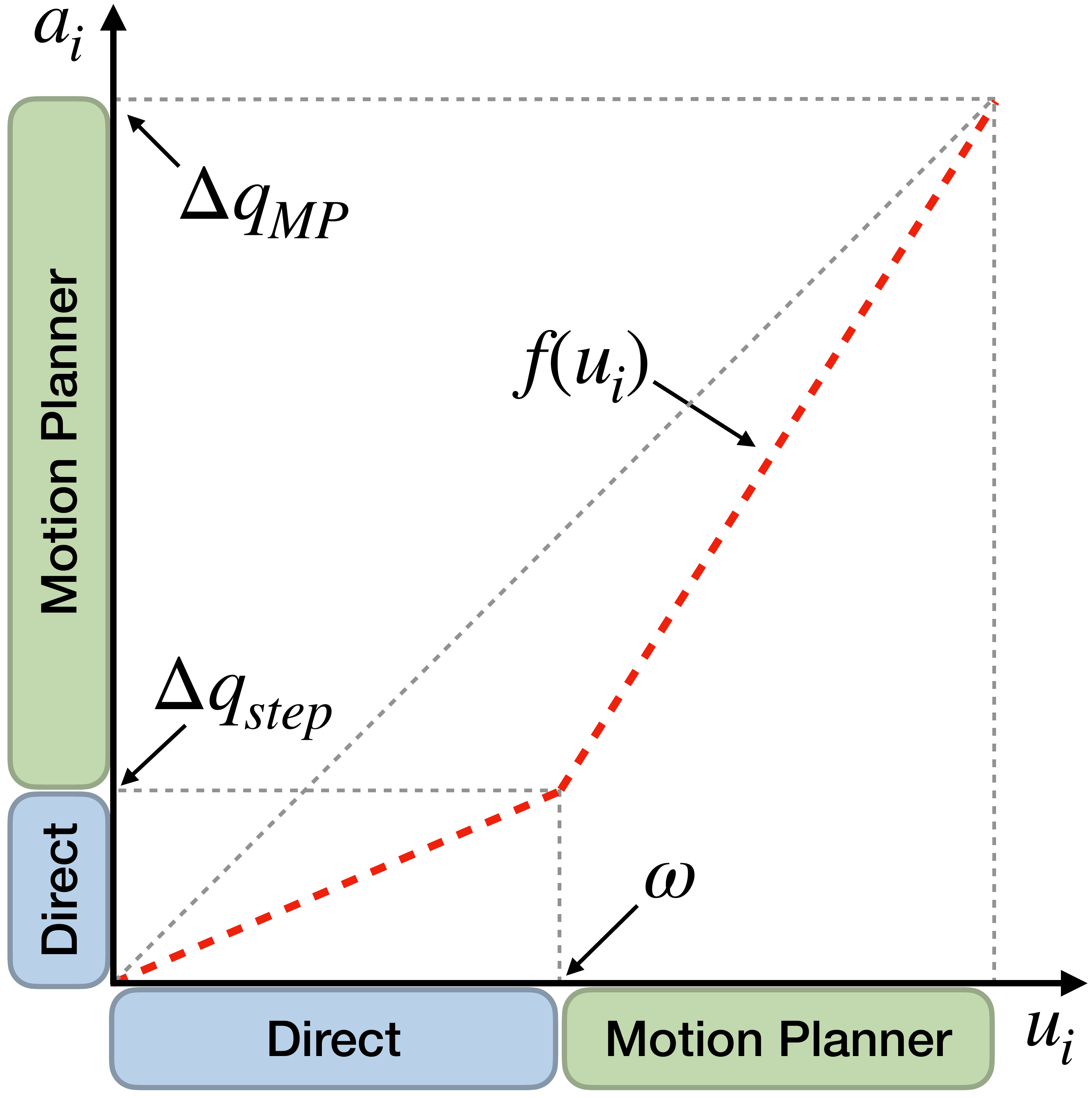}
    \caption{
        To balance the chance of choosing direct action execution and motion planning during exploration, we increase the action space for direct action execution $\mathcal{A}$.
    }
    \vspace{-20pt}
    \label{fig:rescaling}
\end{wrapfigure}

The proposed motion planner augmented action space $\tilde{\mathcal{A}}=[-\Delta q_\text{MP},\Delta q_\text{MP}]^d$ extends the typical action space for model-free RL, $\mathcal{A}=[-\Delta q_\text{step},\Delta q_\text{step}]^d$.
An action $\tilde{a}$ from the original action space $\mathcal{A}$ is directly executed with a feedback controller. On the other hand, an action from outside of $\mathcal{A}$ is handled by the motion planner. However, in practice, $\Delta q_\text{MP}$ is much larger than $\Delta q_\text{step}$, which results in a drastic difference between the proportions of the action spaces for direct action execution and motion planning. Especially with high-dimensional action spaces, this leads to very low probability $\left({\Delta q_\text{step}}/{\Delta q_\text{MP}}\right)^{d}$
of selecting direct action execution during exploration. Hence, this naive action space partitioning biases using motion planning over direct action execution and leads to failures of learning contact-rich manipulation tasks.

To circumvent this issue, we balance the ratio of sampling actions for direct action execution $\tilde{a} \in \mathcal{A}$ and motion plan actions $\tilde{a} \in \tilde{\mathcal{A}} \setminus \mathcal{A}$ by rescaling the action space. \myfigref{fig:rescaling} illustrates the distribution of direct action and motion plan action in $\tilde{\mathcal{A}}$ (y-axis) and the desired distribution (x-axis). To increase the portion of direct action execution, we apply a piecewise linear function~$f$ to the policy output $u \in [-1,1]^{d}$ and get joint displacements $\Delta q$, as shown by the red line in \myfigref{fig:rescaling}.
From the policy output $u$, the action (joint displacement) of the $i$-th joint can be computed by
\begin{equation}
\label{eq:action_rescaling}
\tilde{a}_i=f(u_i) = \begin{cases}
			\frac{\Delta q_\text{step}}{\omega}u_i & \text{$|u_i| \leq \omega$} \\
            \text{sign}(u_i)\left[\Delta q_\text{step} + (\Delta q_\text{MP}-\Delta q_\text{step})\left(\frac{|u_i|-\omega}{1 - \omega}\right)\right] & \text{otherwise}
		 \end{cases},
\end{equation}
where $\omega\in[0,1]$ determines the desired ratio between the sizes of the two action spaces and $\text{sign}(\cdot)$ is the sign function.

\subsection{Training Details}
\label{sec:training_details}

We model the policy $\pi_\phi$ as a neural network. The policy and critic networks consist of 3 fully connected layers of 256 hidden units with ReLU nonlinearities.
The policy outputs the mean and standard deviation of a Gaussian distribution over an action space. To bound the policy output $u$ in $[-1, 1]$, we apply $\tanh$ activation to the policy output.
Before executing the action in the environment, we transform the policy output with the action rescaling function $f(u)$ described in \myeqref{eq:action_rescaling}.
The policy is trained using a model-free RL method, Soft Actor-Critic~\citep{haarnoja2018sac}. 

To improve sample efficiency, we randomly sample $M$ intermediate transitions of a path from the motion planner, and store the sampled transitions in the replay buffer. By making use of these additional transitions, the agent experience can cover a wider region in the state space during training (see \mysecref{supp:reuse}). For hyperparameters and more details about training, please refer to \mysecref{supp:training_details}.

\subsection{Motion Planner Implementation}
\label{sec:training}

Our method seamlessly integrates model-free RL and MP through the augmented action space. Our method is agnostic to the choice of MP algorithm. Specifically, we use RRT-Connect~\citep{kuffner2000rrt_connect} from the open motion planning library (OMPL)~\citep{ompl} due to its fast computation time.
After the motion planning, the resulting path is smoothed using a shortcutting algorithm~\citep{geraerts2007creating}.
For collision checking, we use the collision checking function provided by the MuJoCo physics engine~\citep{todorov2012mujoco}.

The expensive computation performed by the motion planner can be a major bottleneck for training. 
Thus, we design an efficient MP procedure with several features.
First, we reduce the number of costly MP executions by using a simpler motion planner that attempts to linearly interpolate between the initial and goal states instead of the sampling-based motion planner.
If the interpolated path is collision-free, our method uses this path for execution and skips calling the expensive MP algorithm. If the path has collision, then RRT-Connect is used to find a collision-free path amongst obstacles. 

Moreover, the RL policy can predict a goal joint state that is in collision or not reachable. A simple way to resolve it is to ignore the current action and sample a new action. However, it slows down training because the policy can repeatedly output invalid actions, especially in an environment with many obstacles. Thus, our method finds an alternative collision-free goal joint state by iteratively reducing the action magnitude and checking collision, similar to \citet{zhang2008efficient}. This strategy prevents the policy from being stuck or wasting samples, which results in improved training efficiency.
Finally, we allow the motion planner to derive plans while grasping an object by considering the object as a part of the robot once it holds the object. 

\section{Experiments}

We design our experimental evaluation to answer the following questions: (1)~Can MoPA-RL solve complex manipulation tasks in obstructed environments more efficiently than conventional RL algorithms? (2)~Is MoPA-RL better able to explore the environment? (3)~Does MoPA-RL learn policies that are safer to execute? 

\subsection{Environments}
To answer these questions, we conduct experiments on the following hard-exploration tasks in obstructed settings, simulated using the MuJoCo physics engine~\citep{todorov2012mujoco} (see \myfigref{fig:envs} for visualizations):

\begin{itemize}[leftmargin=10mm, nolistsep, itemsep=0.5em]
\item \textbf{2D Push:} A 4-joint 2D reacher needs to push an object into a goal location in the presence of multiple obstacles. 

\item \textbf{Sawyer Push:} A Rethink Sawyer robot arm with 7 DoF needs to push an object into a goal position, both of which are inside a box. 

\item \textbf{Sawyer Lift:} The Sawyer robot arm needs to grasp and lift a block out of a deep box. 

\item \textbf{Sawyer Assembly:} The Sawyer arm needs to move a leg attached to its gripper towards a mounting location while avoiding other legs, and assemble the table by inserting the leg into the hole. 
The environment is built upon the IKEA furniture assembly environment~\citep{lee2019ikea}. 
\end{itemize}

We train \textit{2D Push}, \textit{Sawyer Push} and \textit{Sawyer Assembly} using sparse rewards: when close to the object or goal the agent receives a reward proportional to the distance between end-effector and object or object and goal; otherwise there is no reward signal. For \textit{Sawyer Lift}, we use a shaped reward function, similar to \citet{fan2018surreal}. In all tasks, the agent receives a sparse completion reward upon solving the tasks. Further details about the environments can be found in \mysecref{supp:environment_details}.

\subsection{Baselines}

We compare the performance of our method with the following approaches:

\begin{itemize}[leftmargin=10mm, nolistsep, itemsep=0.5em]
\item \textbf{SAC:} A policy trained to predict displacements in the robot's joint angles using Soft Actor-Critic (SAC, \cite{haarnoja2018sac}), a state-of-the-art model-free RL algorithm.

\item \textbf{SAC Large:}  A variant of SAC predicts joint displacements in the extended action space $\tilde{\mathcal{A}}$. To realize a large joint displacement, a large action $\tilde{a}$ is linearly interpolated into a sequence of small actions $\{a_{1}, \dots a_{m}\}$ such that $\tilde{a} = \sum_{i=1}^{m}a_i$ and $||a_{i}||_\infty \leq \Delta q_\text{step}$. This baseline tests the importance of collision-free MP for a large action space.

\item \textbf{SAC IK:} A policy trained with SAC which outputs displacement in Cartesian space instead of the joint space. For 3D manipulation tasks, the policy also outputs the displacement in end-effector orientation as a quaternion. Given the policy output, a target \emph{joint} state is computed using inverse kinematics and the joint displacement is applied to the robot.

\item \textbf{MoPA-SAC (Ours):} Our method predicts a joint displacement in the extended action space.

\item \textbf{MoPA-SAC Discrete:} Our method with an additional discrete output that explicitly chooses between MP and RL, which replaces the need for a threshold $\omega$. 

\item \textbf{MoPA-SAC IK:} Our method with end-effector space control instead of joint space displacements. Again, inverse kinematics is used to obtain a target joint state, which is either directly executed or planned towards with the motion planner. 
\end{itemize}

\begin{figure*}[t!]
\begin{minipage}{\textwidth}
  \centering
  \begin{subfigure}[t]{0.24\textwidth}
      \includegraphics[width=\textwidth]{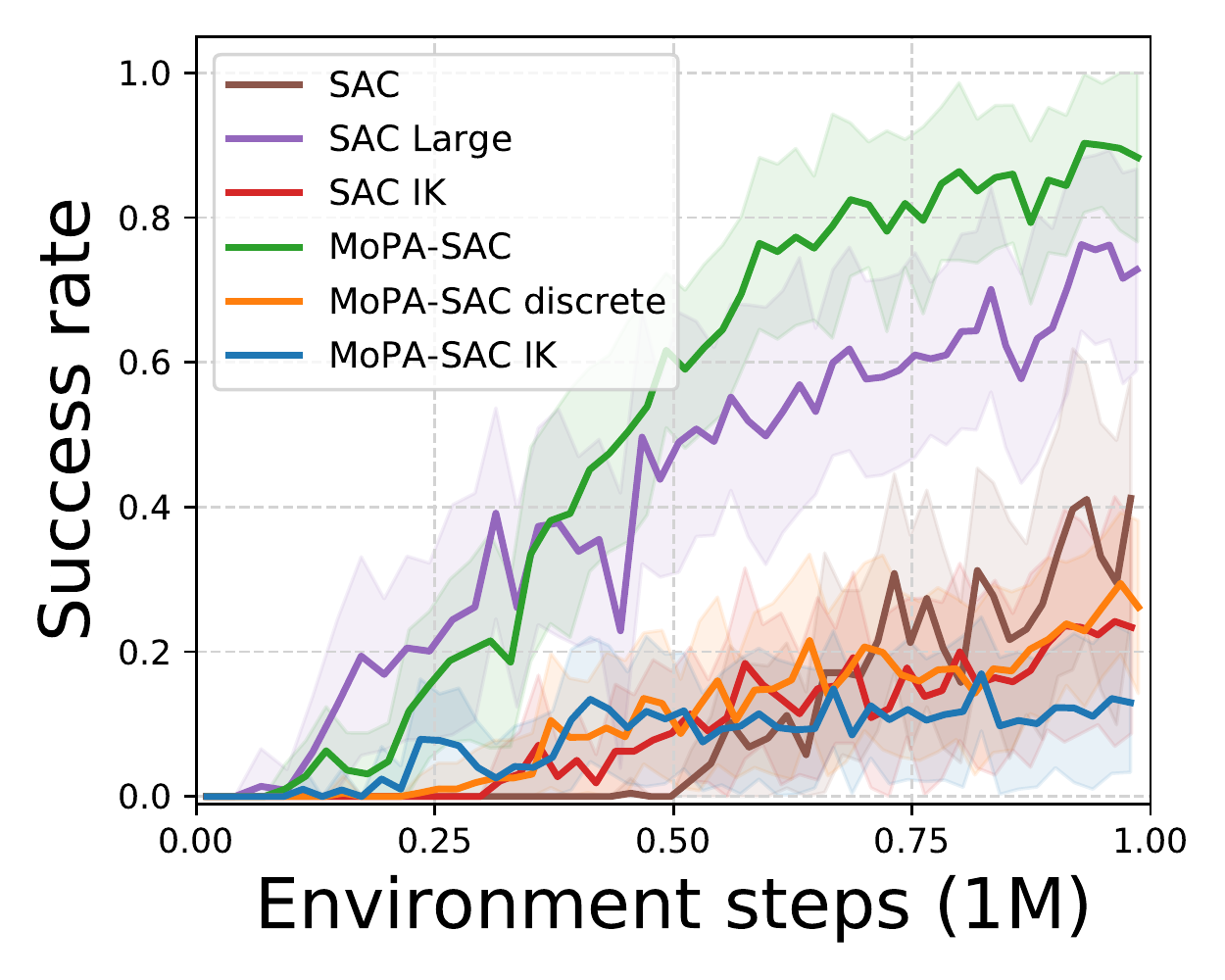}
      \caption{2D Push}
  \end{subfigure}
  \begin{subfigure}[t]{0.24\textwidth}
      \includegraphics[width=\textwidth]{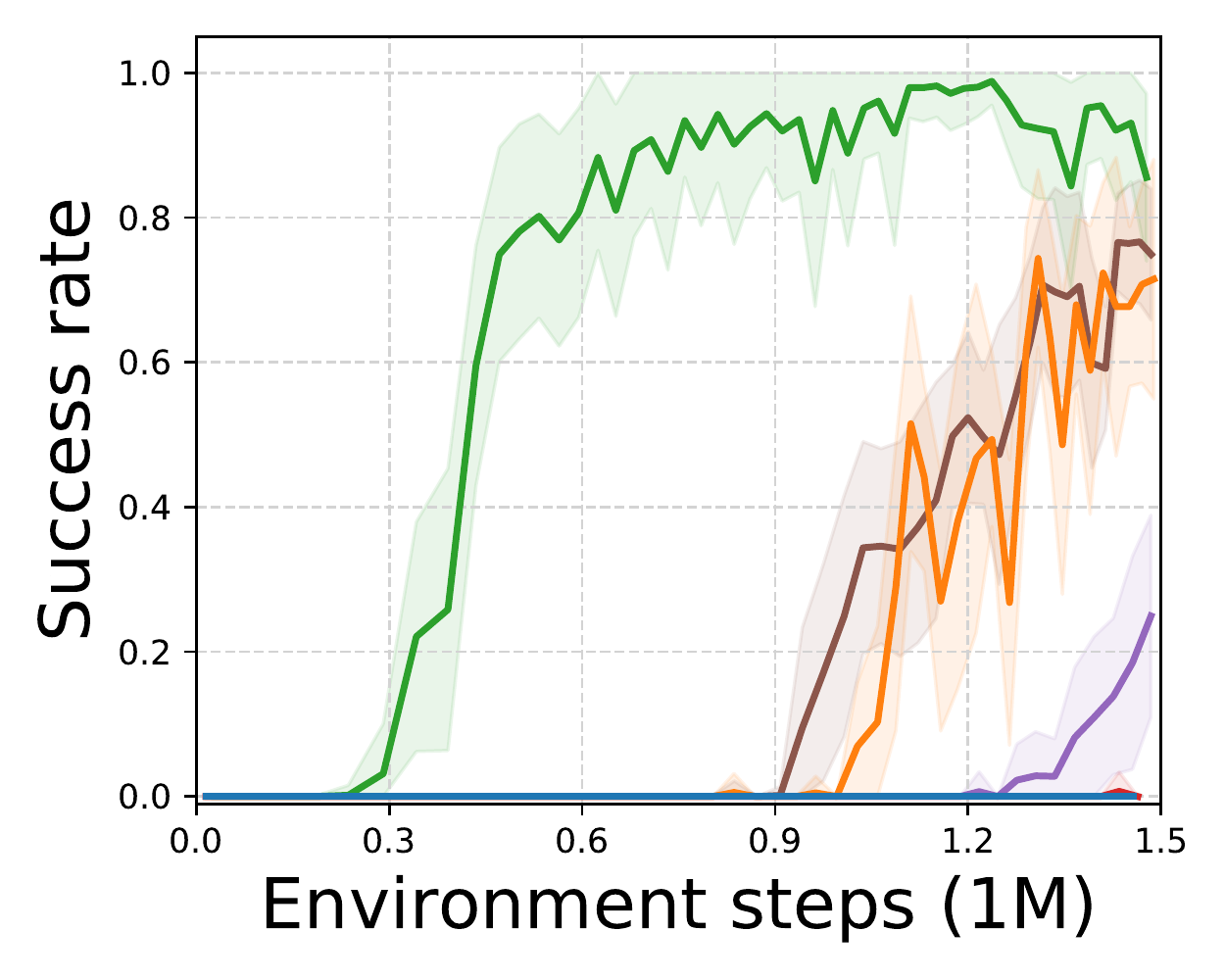}
      \caption{Sawyer Push}
  \end{subfigure}
  \begin{subfigure}[t]{0.24\textwidth}
      \includegraphics[width=\textwidth]{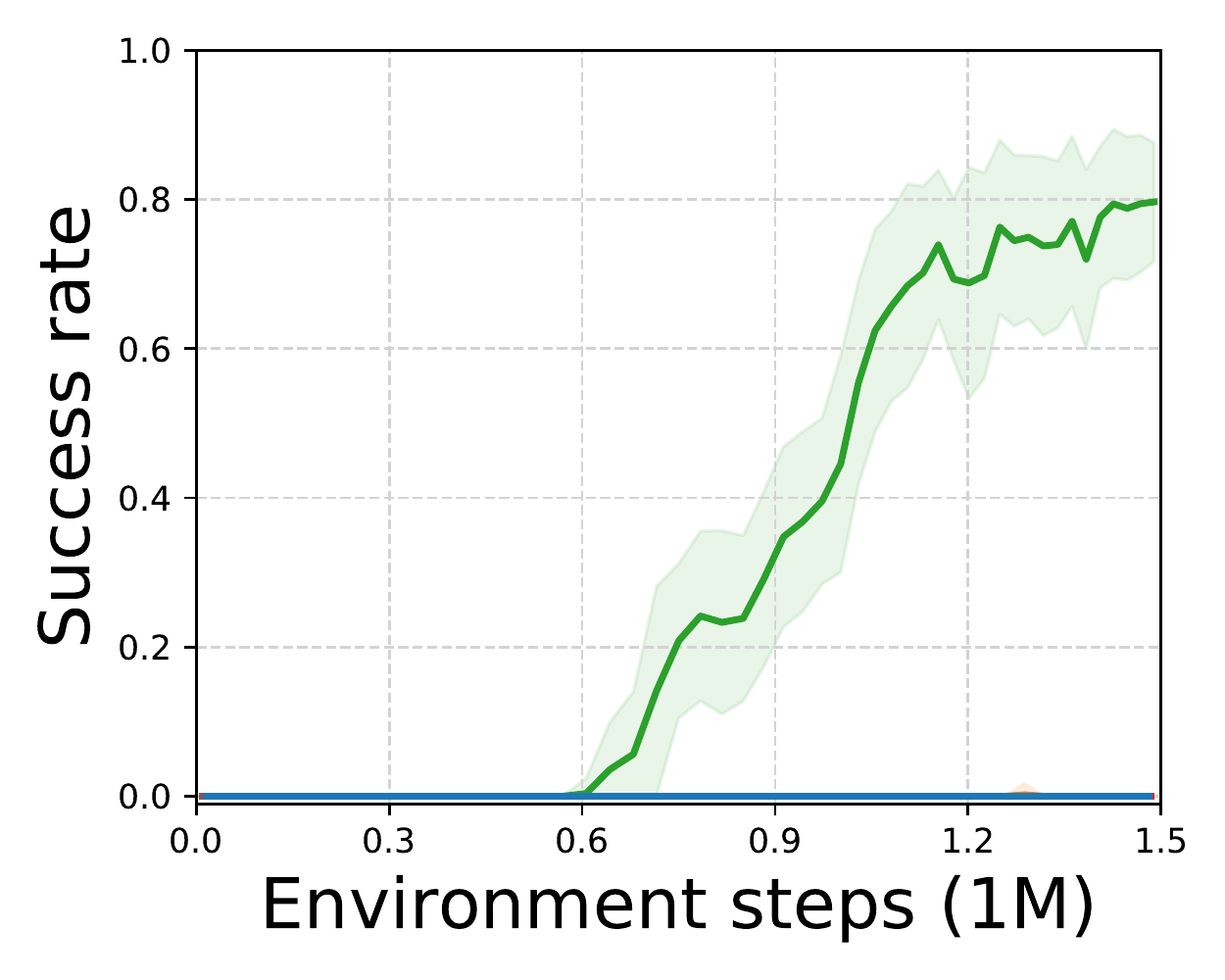}
      \caption{Sawyer Lift}
  \end{subfigure}
  \begin{subfigure}[t]{0.24\textwidth}
    \includegraphics[width=\textwidth]{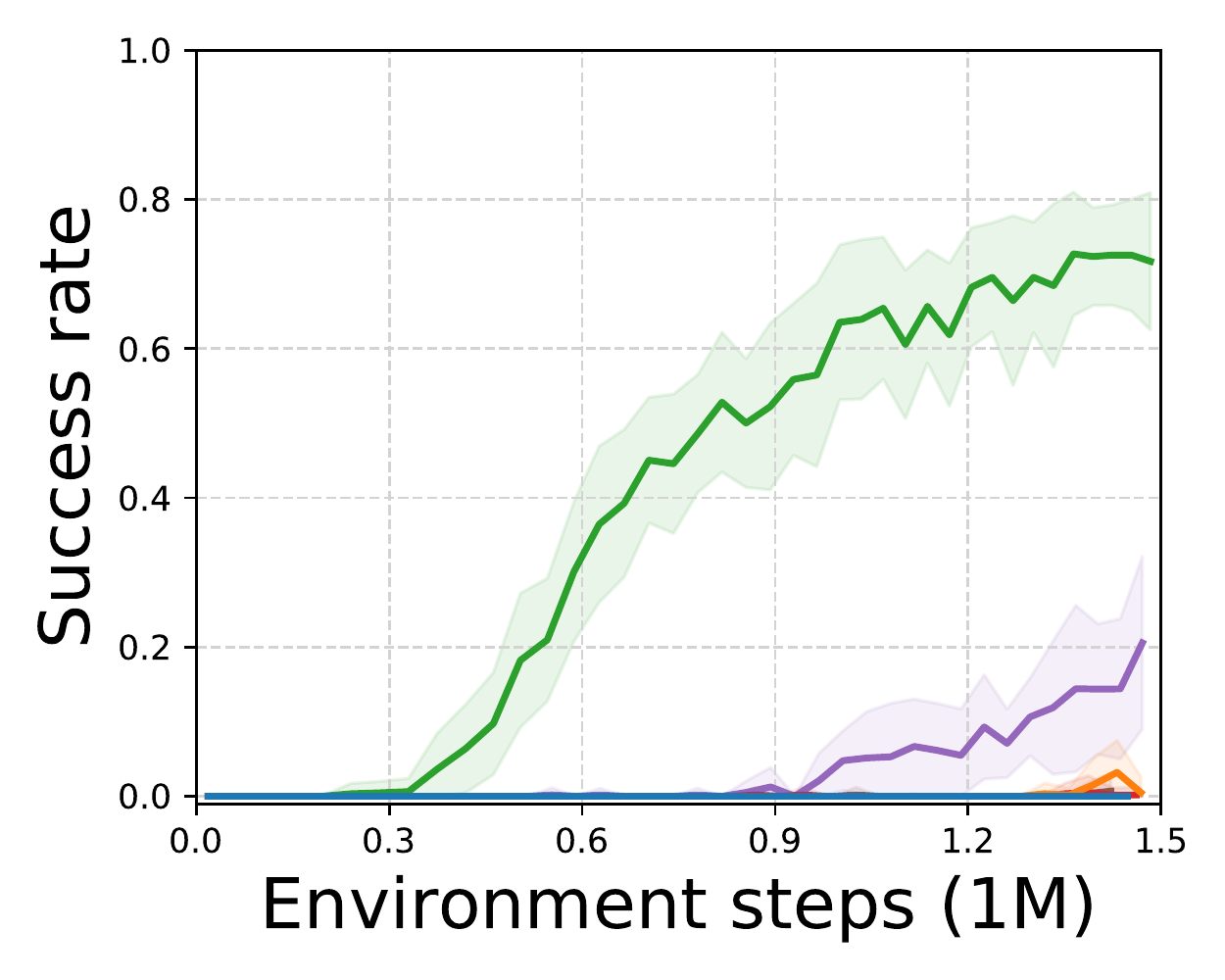}
      \caption{Sawyer Assembly}
  \end{subfigure}
\end{minipage}
\caption{
Success rates of our MoPA-SAC (green) and baselines averaged over 4 seeds. All methods are trained for the same number of environment steps. MoPA-SAC can solve all four tasks by leveraging the motion planner and learn faster than the baselines. 
}
\vspace{-15pt}
\label{fig:result}
\end{figure*}

\subsection{Efficient RL with Motion Planner Augmented Action Spaces}
\label{sec:results}

We compare the learning performance of all approaches on four tasks in \myfigref{fig:result}. Only our MoPA-SAC is able to learn all four tasks, while other methods converge more slowly or struggle to obtain any rewards. The difference is especially large in the \textit{Sawyer Lift} and \textit{Sawyer Assembly} tasks. This is because precise movements are required to maneuver the robot arm to reach inside the box that surrounds the objects or avoid the other table legs while moving the leg in the gripper to the hole. While conventional model-free RL agents struggle to learn such complex motions from scratch, our approach can leverage the capabilities of the motion planner to successfully learn to produce collision-free movements.

\begin{wrapfigure}{r}{0.45\textwidth}
    \vspace*{-0.3in}
    \centering
    \includegraphics[width=0.9\linewidth]{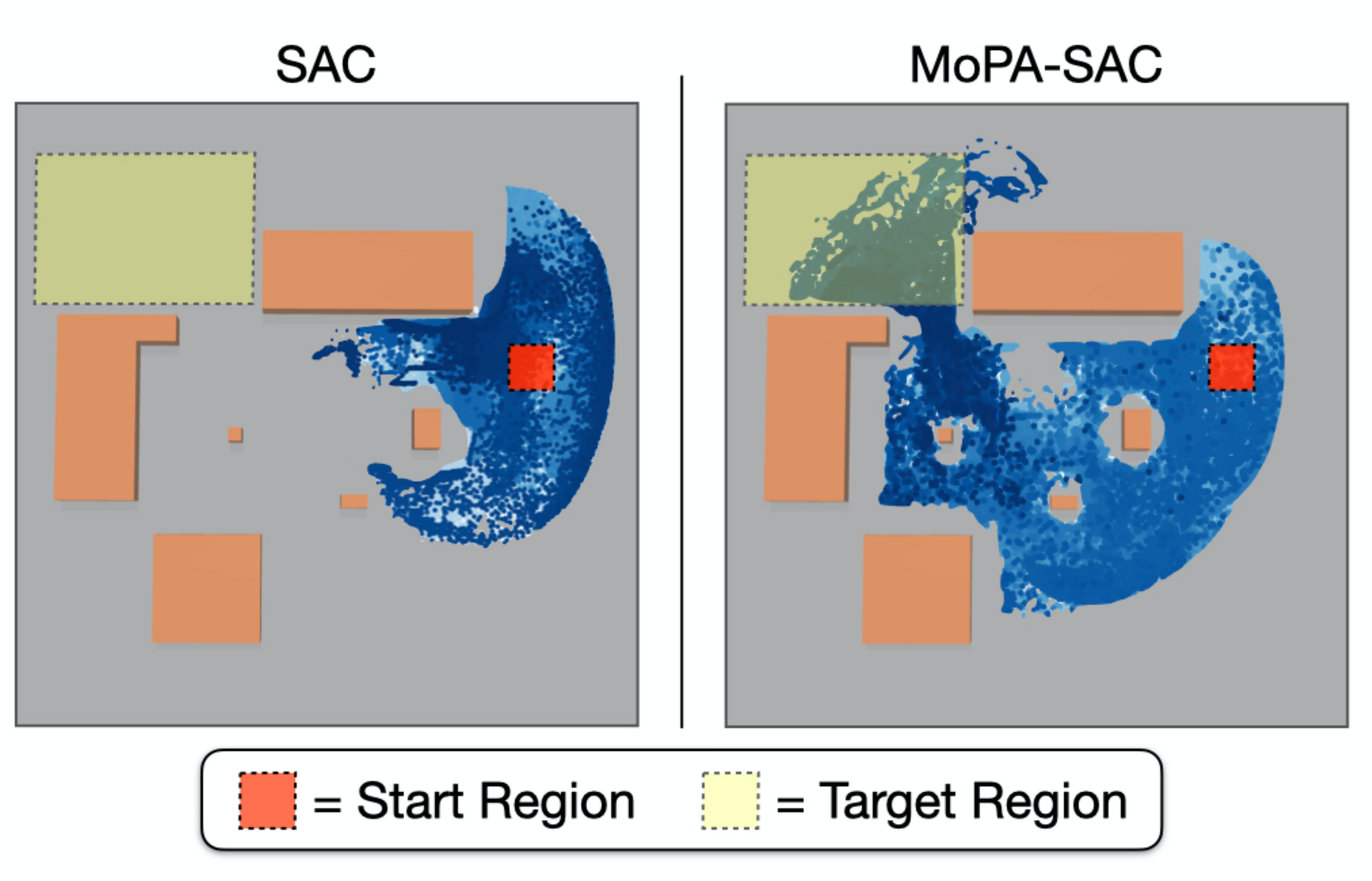}
    \caption{End-effector positions of SAC (\textbf{left}) and MoPA-SAC (\textbf{right}) after the first 100k training environment steps in \emph{2D Push} are plotted in blue dots. The use of motion planning allows the agent to explore the environment more widely early on in training.}
    \label{fig:exploration}
    \vspace*{-0.1in}
\end{wrapfigure}

To further analyze \emph{why} augmenting RL agents with MP capability improves learning performance, we compare the exploration behavior in the first 100k training steps of our MoPA-SAC agent and the conventional SAC agent on the \textit{2D Push} task in \myfigref{fig:exploration}. The SAC agent initially explores only in close proximity to its starting position as it struggles to find valid trajectories between the obstacles. In contrast, the motion planner augmented agent explores a wider range of states by using MP to find collision-free trajectories to faraway goal states. This allows the agent to quickly learn the task, especially in the presence of many obstacles. Efficient exploration is even more challenging in the obstructed 3D environments. Therefore, only the MoPA-SAC agent that leverages the motion planner for efficient exploration is able to learn the manipulation tasks.

The comparison between different action spaces for our method in \myfigref{fig:result} shows that directly predicting joint angles and using the motion planner based on action magnitude leads to the best learning performance (\textit{MoPA-SAC (Ours)}). In contrast, computing the target joint angles for the motion planner using inverse kinematics (\textit{MoPA-SAC IK}) often produces configurations that are in collision with the environment, especially when manipulations need to be performed in narrow spaces. \textit{MoPA-SAC~Discrete} needs to jointly learn how and \emph{when} to use MP by predicting a discrete switching variable. We find that this approach rarely uses MP, leading to worse performance.

\subsection{Safe Policy Execution}

The ability to execute safe collision-free trajectories in obstructed environments is important for the application of RL in the real world. We hypothesize that the MoPA-RL agents can leverage MP to learn trajectories that avoid unnecessary collisions. To validate this, we report the average contact force of all robot joints on successful rollouts from the trained policies in \myfigref{fig:safety}. The MoPA-RL agents show low average contact forces that are mainly the result of the necessary contacts with the objects that need to be pushed or lifted. Crucially, these agents are able to perform the manipulations \emph{safely} while avoiding collisions with obstacles. In contrast, conventional RL agents are unable to effectively avoid collisions in the obstructed environments, leading to high average contact forces.

\subsection{Ablation Studies}

\begin{figure*}[t!]
\begin{minipage}{\textwidth}
  \centering
  \begin{subfigure}[t]{0.30\textwidth}
    \includegraphics[width=\textwidth]{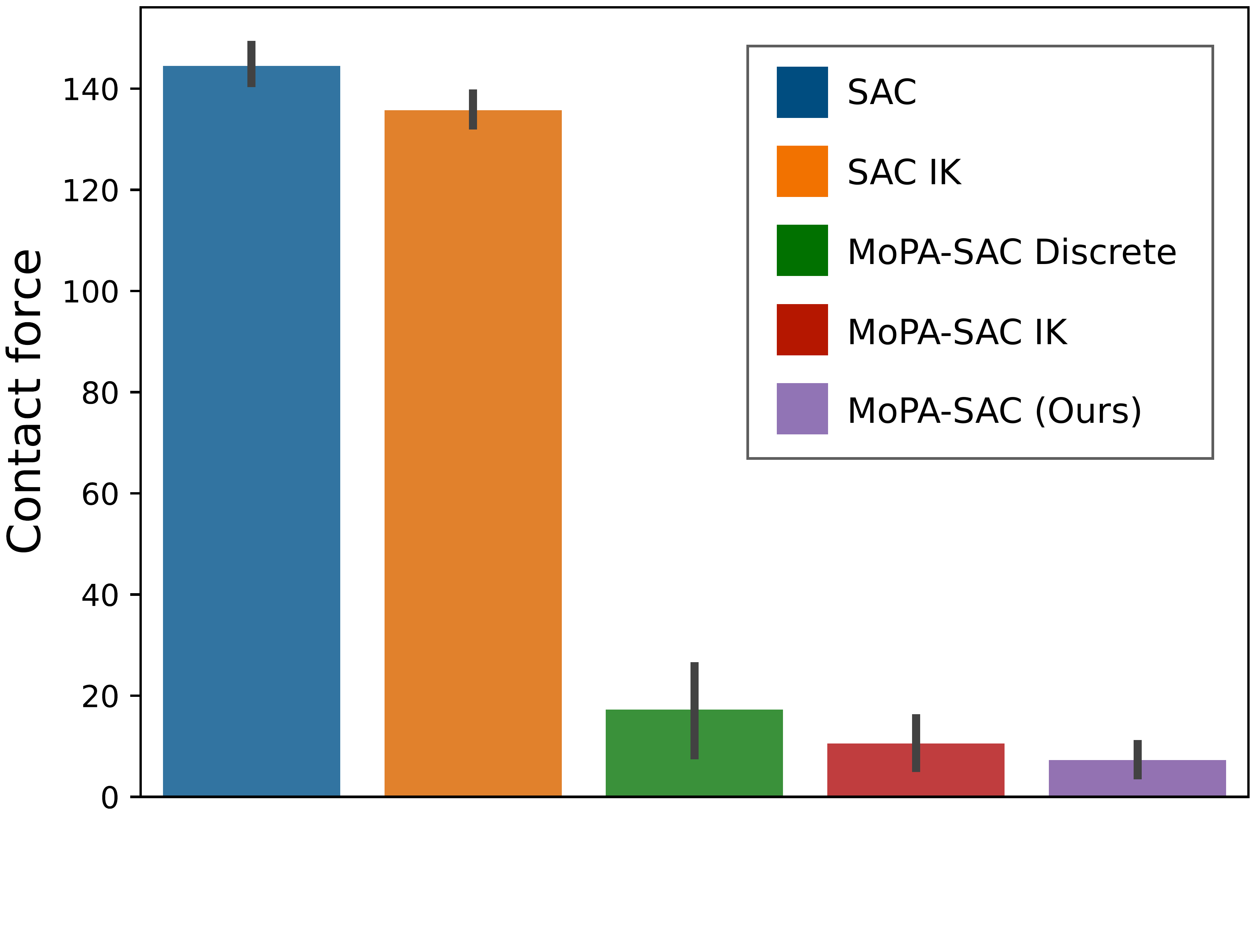}    
    \caption{Contact force}
    \label{fig:safety}
  \end{subfigure}
  \begin{subfigure}[t]{0.29\textwidth}  
      \includegraphics[width=\textwidth]{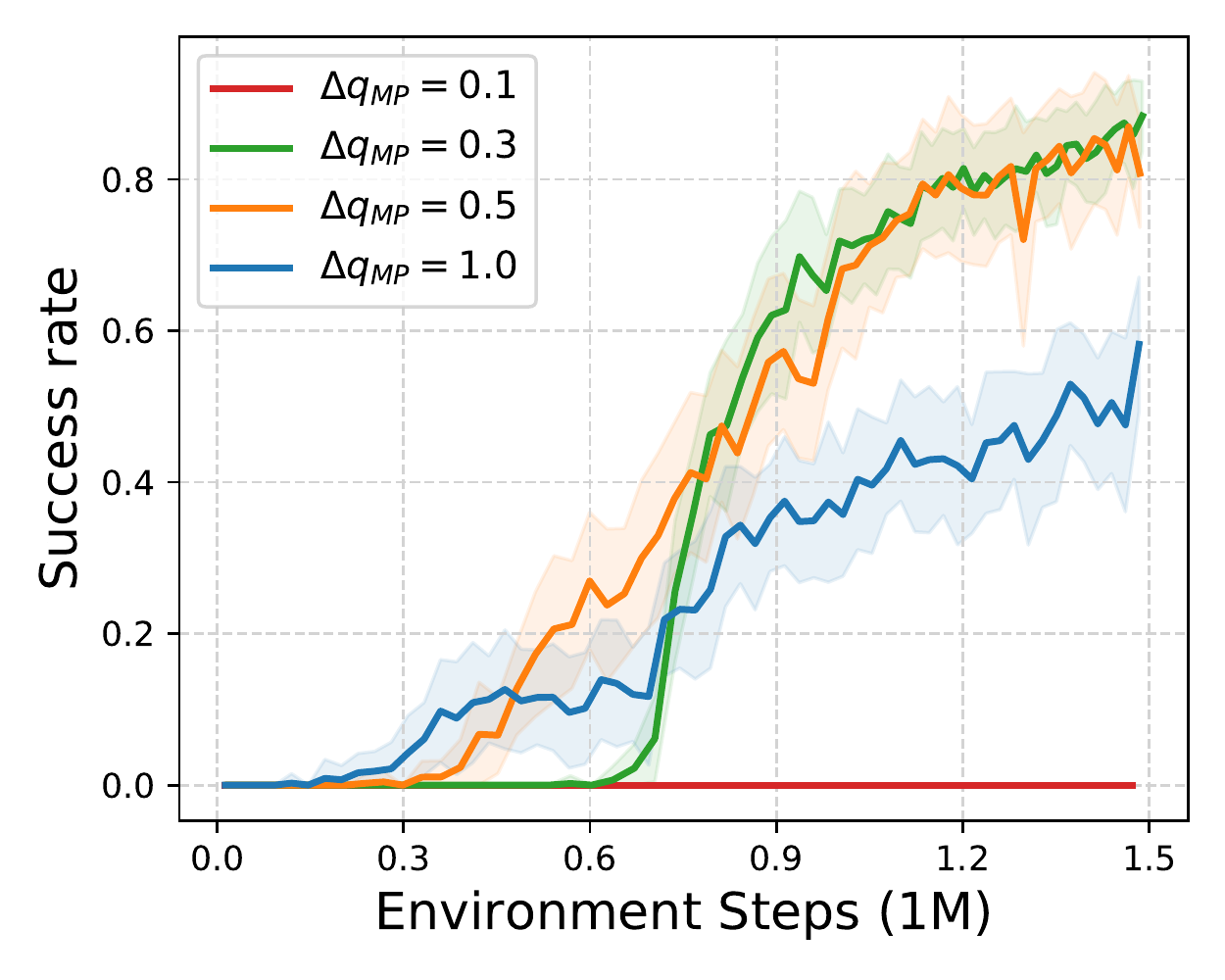}
      \caption{Action range $\Delta q_\text{MP}$}
      \label{fig:ablation1}
  \end{subfigure}
  \begin{subfigure}[t]{0.29\textwidth}
      \includegraphics[width=\textwidth]{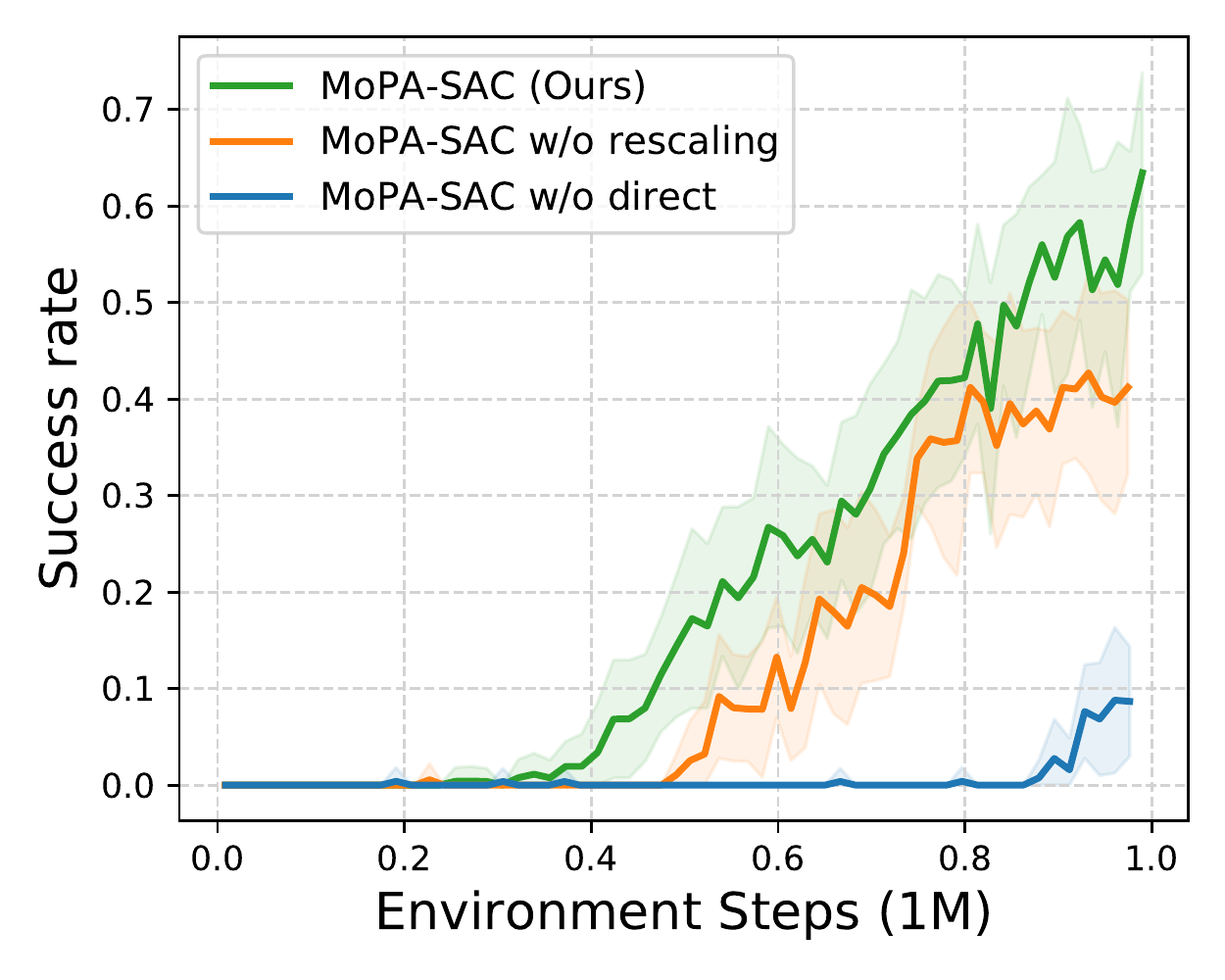}
      \caption{Action space rescaling}
      \label{fig:ablation2}
  \end{subfigure}
\end{minipage}
\caption{(a) Averaged contact force in an episode over 7 executions in \emph{2D Push}. Leveraging a motion planner, all variants of our method naturally learn collision-safe trajectories. (b) Comparison of our model with different action range values $\Delta q_\text{MP}$ on \textit{Sawyer Lift}. (c) Comparison of our model w/ and w/o action rescaling or w/o direct action execution on \textit{Sawyer Lift}. 
}
\vspace{-1em}
\end{figure*} 

\paragraph{Action range:} We analyze the influence of the action range $\Delta q_\text{MP}$ on task performance in \myfigref{fig:ablation1}. We find that for too small action ranges the policy cannot efficiently explore the environment and does not learn the task. Yet, for too large action ranges the number of possible actions the agent needs to explore is large, leading to slow convergence. In between, our approach is robust to the choice of action range and able to learn the task efficiently. 

\paragraph{Action rescaling and direct action execution:} In \myfigref{fig:ablation2} we ablate the action space rescaling introduced in \mysecref{sec:rescaling}.
We find that action space rescaling improves learning performance by encouraging balanced exploration of both single-step and motion planner action spaces. More crucial is however our hybrid action space formulation with direct and MP action execution: MoPA-SAC trained \emph{without} direct action execution struggles on contact-rich tasks, since it is challenging to use the motion planner for solving contact-rich object manipulations.

\section{Conclusion} 
\label{sec:conclusion}

In this work, we propose a flexible framework that combines the benefits of both motion planning and reinforcement learning for sample-efficient learning of continuous robot control in obstructed environments. Specifically, we augment a model-free RL with a sampling-based motion planner with minimal task-specific knowledge. The RL policy can learn when to use the motion planner and when to take a single-step action directly through reward maximization. The experimental results show that our approach improves the training efficiency over conventional RL methods, especially on manipulation tasks in the presence of many obstacles. These results are promising and motivate future work on using more advanced motion planning techniques in the action space of reinforcement learning. Another interesting direction is the transfer of our framework to real robot systems.

\clearpage

{\small
\bibliography{bib/conference,bib/deep_learning,bib/rl,bib/drl,bib/hrl,bib/env,bib/robotics, bib/motion_planning, bib/marl, bib/supplementary}
}

\clearpage
\appendix

\section{Additional Ablation Studies}

We provide further analysis: (1) the effect of reusing motion planning trajectories to augment training data (\mysecref{supp:reuse}); (2) the ablation on the  action space rescaling (\mysecref{supp:rescaling}); (3) the performance of our approach in uncluttered environments compared to baselines (\mysecref{supp:uncluttered}); (4) the ablation of invalid target joint state handling (\mysecref{supp:invalid_handling}); (5) the ablation of motion planner algorithms (\mysecref{supp:diff_motion_planner}); and (6) the ablation of RL algorithms (\mysecref{supp:diff_model-free}).

\subsection{Reuse of Motion Plan Trajectories}
\label{supp:reuse}

As mentioned in \mysecref{sec:training_details}, to improve sample efficiency of motion plan actions, we sample $M$ intermediate trajectories of the motion plan trajectory $\tau_{0:H}=(q_t, q_{t+1}, \dots, q_{t+H})$ and augment the replay buffer with sub-sampled motion plan transitions $(s_{t+a_i}, \Delta \tau_{a_i:b_i}, s_{t+b_i}, \tilde{R}(s_{t+a_i}, \Delta \tau_{a_i:b_i}))$, where $a_i < b_i \in [0, H]$ and $i \in [1, M]$ (see \myalgref{alg:training}).
\myfigref{fig:ablation_reuse} shows the success rates of our model with different $M$, the number of sub-sampled motion plan transitions per motion plan action. 
Reusing trajectory of MP in this way improves the sample efficiency as the success rate starts increasing earlier than the one without reusing motion plan trajectories ($M=0$). However, augmenting too many samples ($M=30, 45$) degrades the performance since it biases the distribution of the replay buffer to motion plan actions and reduces the sample efficiency of the direct action executions, which results in slow learning of contact-rich skills. This biased distribution of transitions leads to convergence towards sub-optimal solutions while the model without bias $M=0$ eventually finds a better solution.

\subsection{Further Study on Action Space Rescaling}
\label{supp:rescaling}

\begin{figure}[h]
    \centering
    \begin{subfigure}[t]{0.4\textwidth}
        \includegraphics[width=\textwidth]{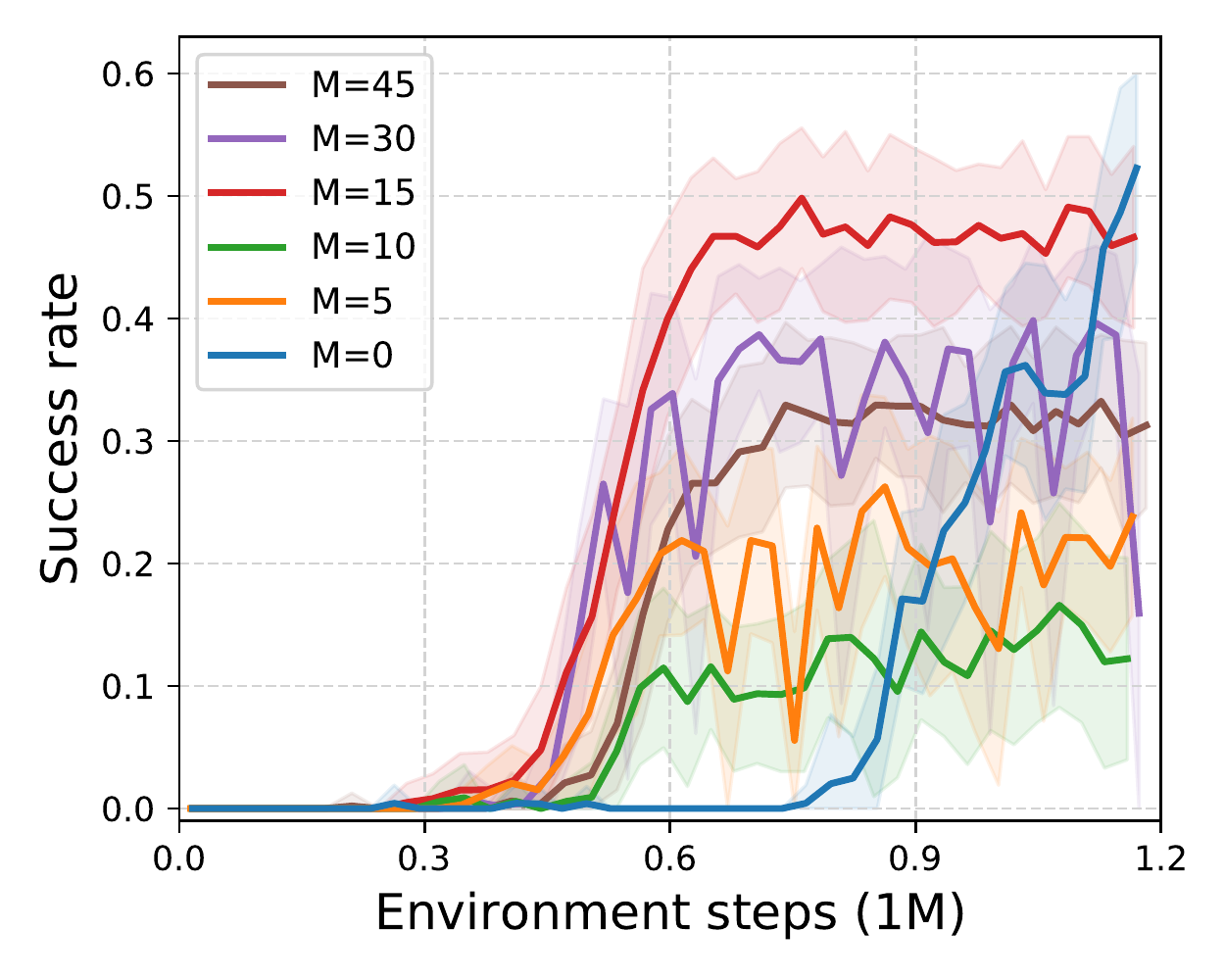}
        \caption{Number of sub-sampled motion plan transitions $M$}
        \label{fig:ablation_reuse}
    \end{subfigure}
    \quad
    \begin{subfigure}[t]{0.4\textwidth}
        \includegraphics[width=\textwidth]{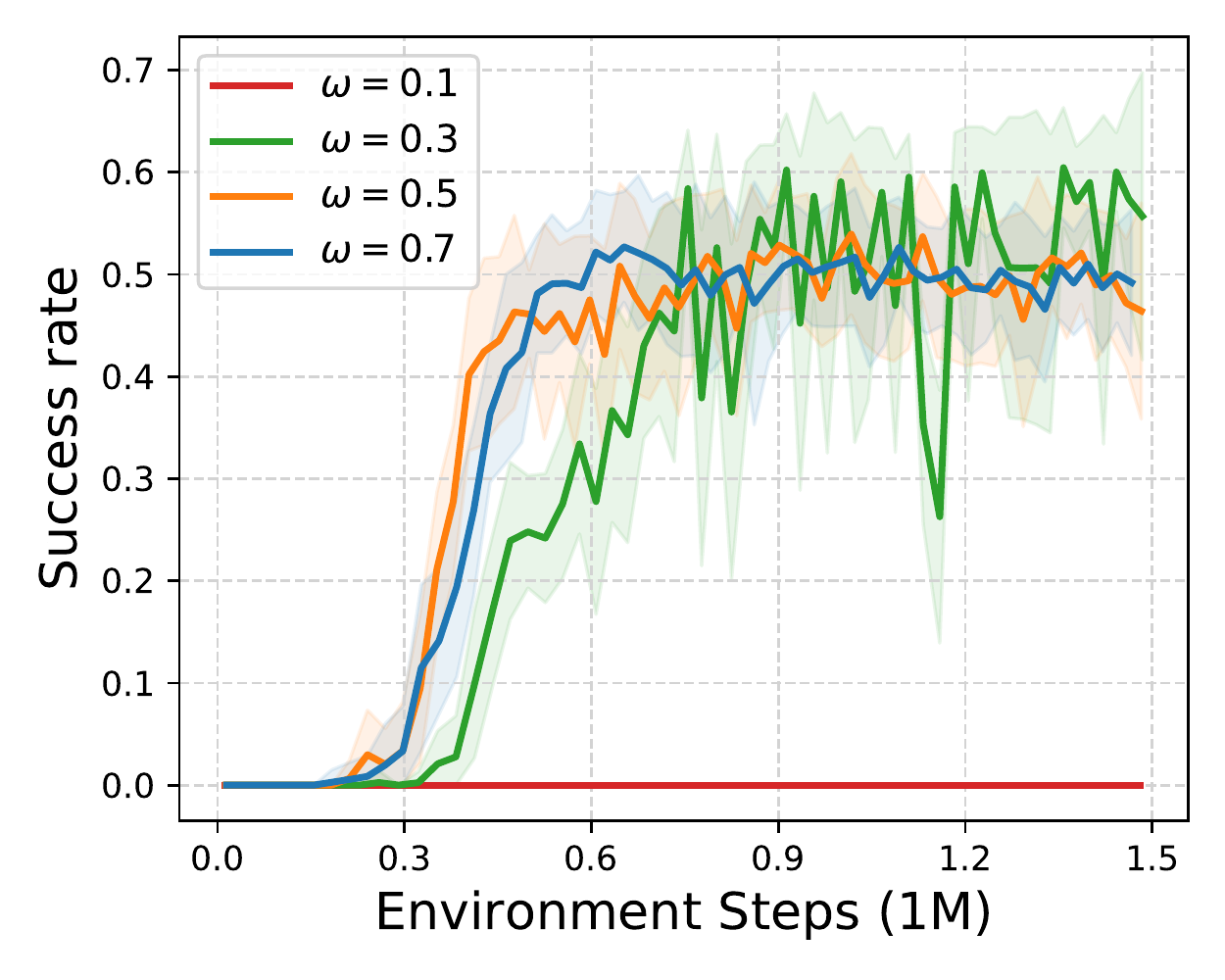}
        \caption{Action space rescaling ratio $\omega$}
        \label{fig:ablation_rescaling}
    \end{subfigure}
    \caption{
        Learning curves of ablated models on \textit{Sawyer Assembly}. 
        (a) Comparison of our MoPA-SAC with different number of samples reused from motion plan trajectories.
        (b) Comparison of our MoPA-SAC with different action space rescaling parameter $\omega$.
    }
\end{figure}

In \mysecref{sec:rescaling}, we propose action space rescaling to balance the sampling ratio between direct action execution and motion planning. As illustrated in \myfigref{fig:ablation_rescaling}, our method without action space rescaling ($\omega=0.1$) fails to solve \textit{Sawyer Assembly} while the policy with action space rescaling learns to solve the task. This failure is mainly because direct action execution is crucial for inserting the table leg and the policy without action space rescaling rarely explores the direct action execution space, which makes the agent struggle to solve the contact-rich task. We also find that the $\omega$ value is not sensitive in \textit{Sawyer Assembly}, as different $\omega$ values achieve similar success rates.

\subsection{Performance in Uncluttered Environments}
\label{supp:uncluttered}

\begin{figure}[h]
  \centering
  \begin{subfigure}[t]{0.4\textwidth}  
      \includegraphics[width=\textwidth]{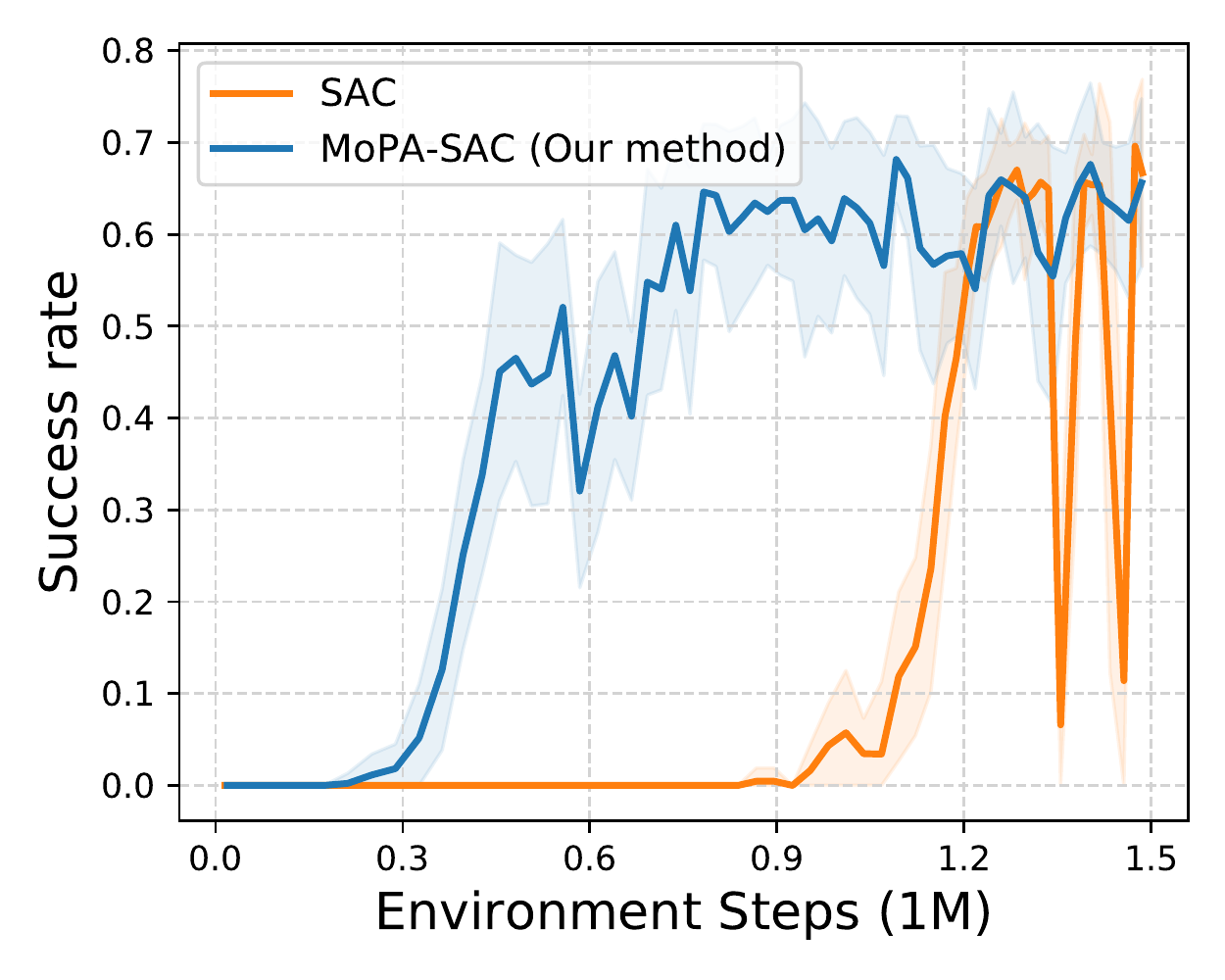}
      \caption{Sawyer Lift w/o box}
      \label{fig:ablation_uncluttered_lift}
  \end{subfigure}
  \quad
  \begin{subfigure}[t]{0.4\textwidth}
      \includegraphics[width=\textwidth]{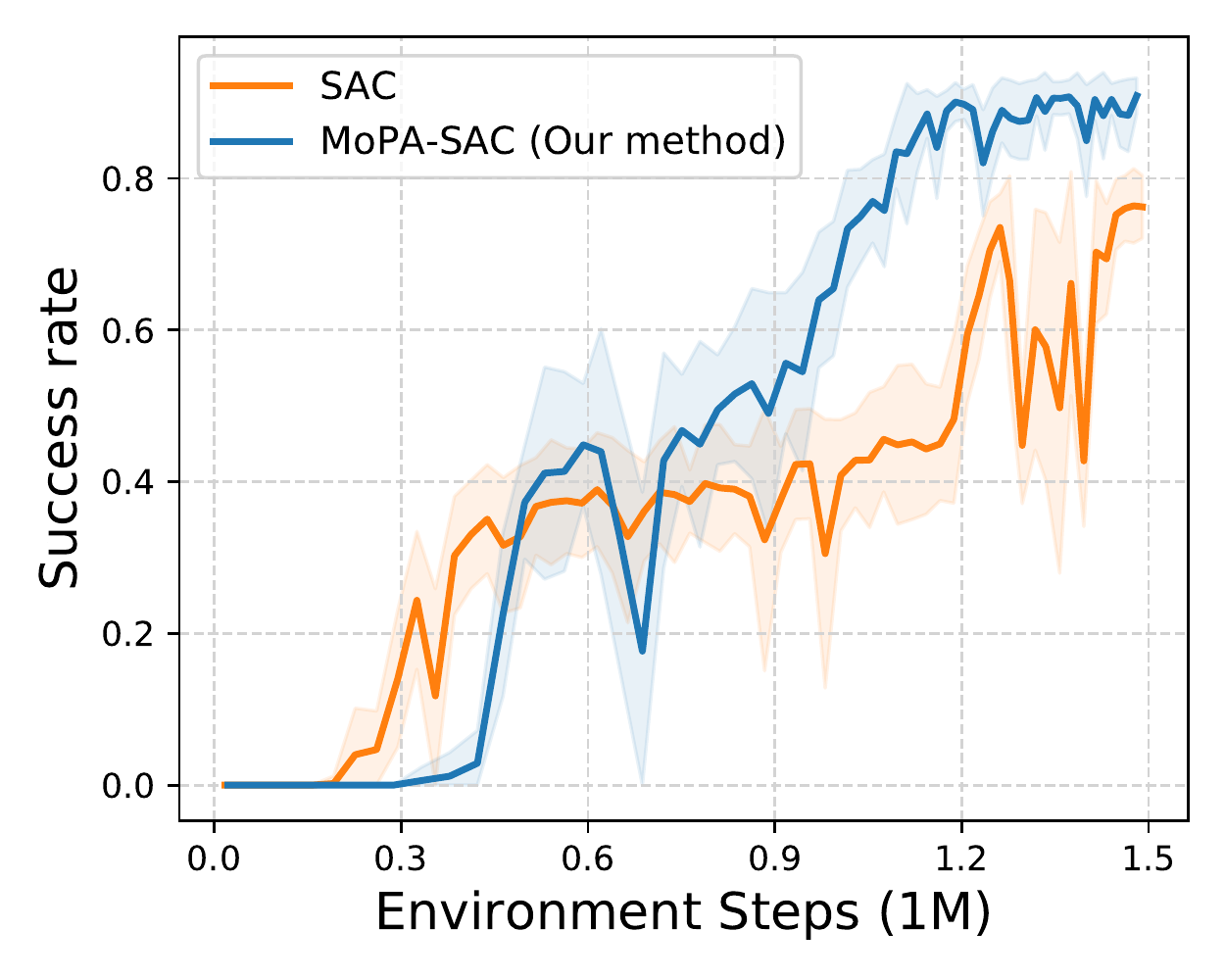}
      \caption{Sawyer Assembly w/o legs}
      \label{fig:ablation_uncluttered_assembly}
  \end{subfigure}
  \caption{
    Success rate on (a) \textit{Sawyer Lift w/o box} and (b) \textit{Sawyer Assembly w/o legs}.
  }
\end{figure}

We further verify whether our method does not degrade the performance of model-free RL in uncluttered environments. Therefore, we remove obstacles, such as a box on a table in \textit{Sawyer Lift} and three other table legs in \textit{Sawyer Assembly}. \myfigref{fig:ablation_uncluttered_lift} and \myfigref{fig:ablation_uncluttered_assembly} show that our method is as sample efficient as the baseline SAC and it is even better in \textit{Sawyer Lift w/o box} because our method does not need to learn how to control an arm for the reaching skill.

\subsection{Handling of Invalid Target Joint States for Motion Planning}
\label{supp:invalid_handling}

When a predicted target joint state $g=q+\tilde{a}$ for motion planning is in collision with obstacles, instead of penalizing or using the invalid action $\tilde{a}$, we search for a valid action by iteratively moving the target joint state towards the current joint state and executing the new valid action, as described in \mysecref{sec:training}. 

\begin{figure}[ht]
  \centering
  \begin{subfigure}[t]{0.4\textwidth}  
      \includegraphics[width=\textwidth]{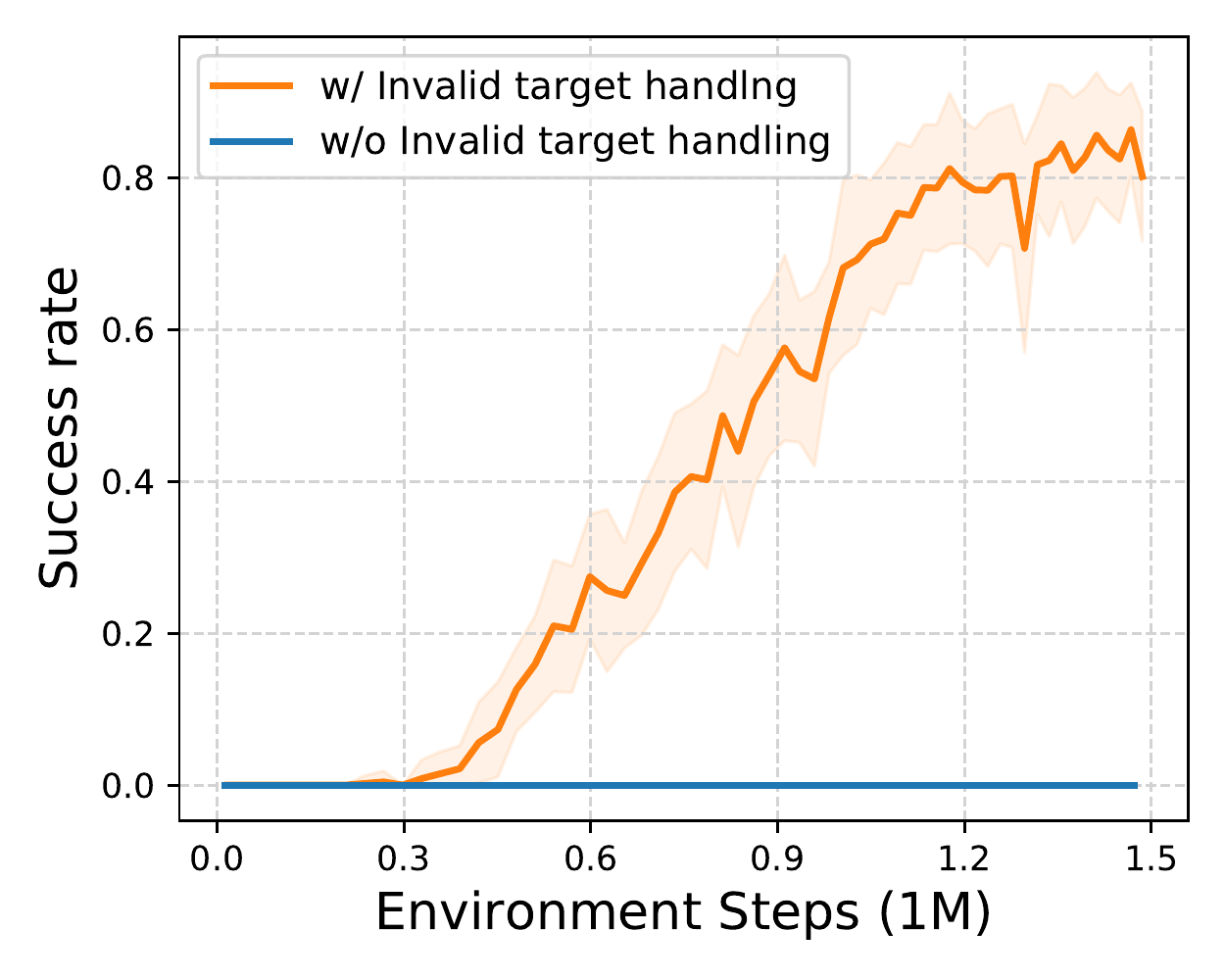}
      \caption{Sawyer Lift}
      \label{fig:ablation_invalid_handling_lift}
  \end{subfigure}
  \quad
  \begin{subfigure}[t]{0.4\textwidth}
      \includegraphics[width=\textwidth]{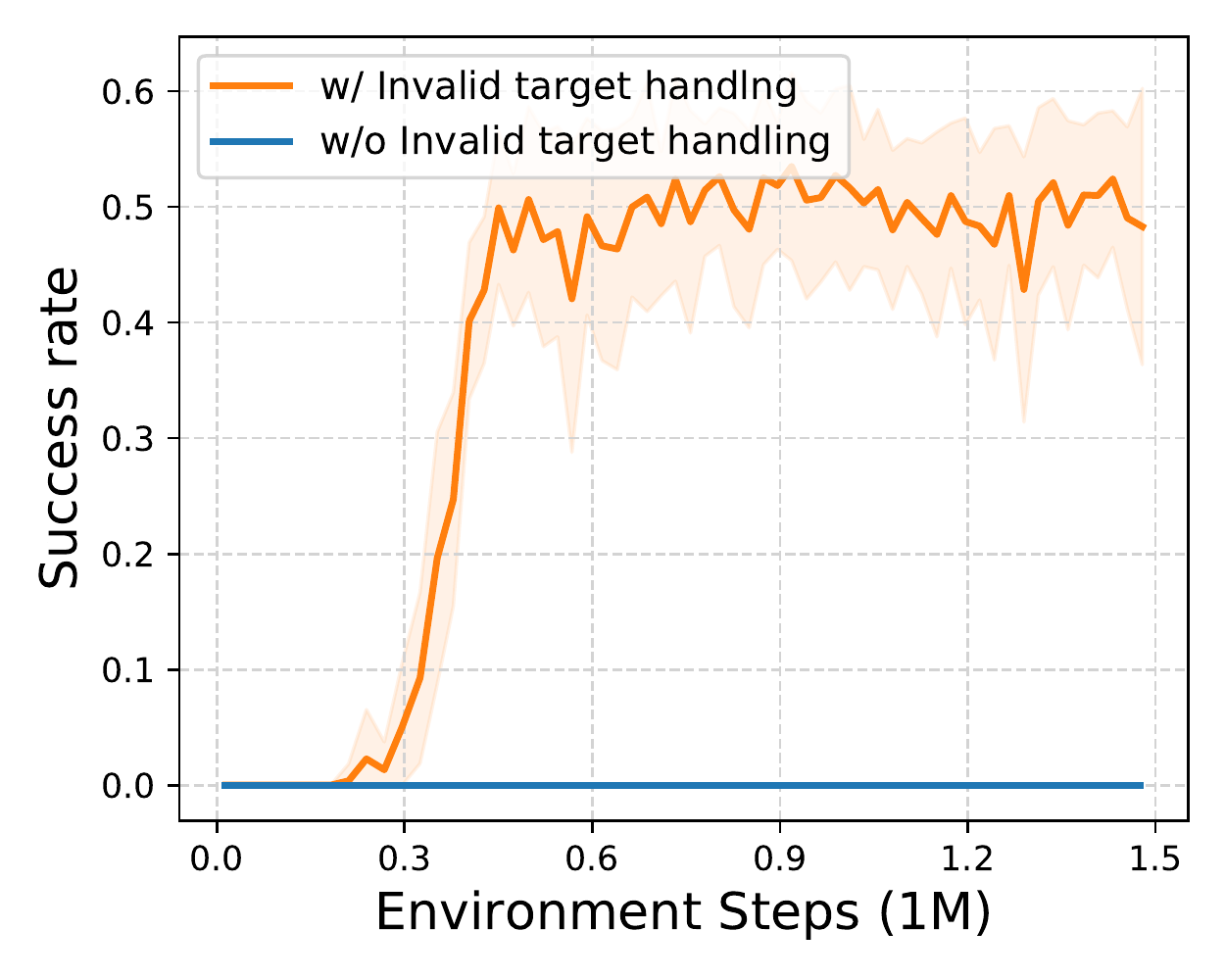}
      \caption{Sawyer Assembly}
      \label{fig:ablation_invalid_handling_assembly}
  \end{subfigure}
  \caption{
    Ablation of invalid target handling on (a) \textit{Sawyer Lift} and (b) \textit{Sawyer Assembly}.
  }
\end{figure}

We investigate the importance of handling the invalid actions for motion planning by comparing to a naive approach for handling invalid actions that the robot does not execute any action and a transition $(s_{t}, a_{t}, r_{t}, s_{t+1})$ is added into a replay buffer, where $s_{t}=s_{t+1}$ and $r_{t}$ is the reward of being at the current state.
\myfigref{fig:ablation_invalid_handling_lift} and \myfigref{fig:ablation_invalid_handling_assembly} show that MoPA-SAC with naive handling of invalid states cannot learn to solve the tasks, which implies that our proposed handling of invalid target state is very crucial to train MoPA-SAC agents. A reason behind this behavior is that the agent can explore the state space even though the invalid target joint state is given.

\subsection{Ablation of Motion Planning Algorithms}
\label{supp:diff_motion_planner}
We test whether our framework is compatible with different motion planning algorithms. \myfigref{fig:ablation_MP} shows the comparison of our method using RRT-Connect and RRT*~\citep{karaman2011RRTStar}. MoPA-SAC with RRT* learns to solve tasks less efficiently than MoPA-SAC with RRT-Connect since, in our experiments, RRT-Connect finds better paths than RRT* within the limited time given to both planners.

\subsection{Ablation of Model-free RL Algorithms}
\label{supp:diff_model-free}

\begin{figure}[ht]
  \centering
  \begin{subfigure}[t]{0.4\textwidth}  
      \includegraphics[width=\textwidth]{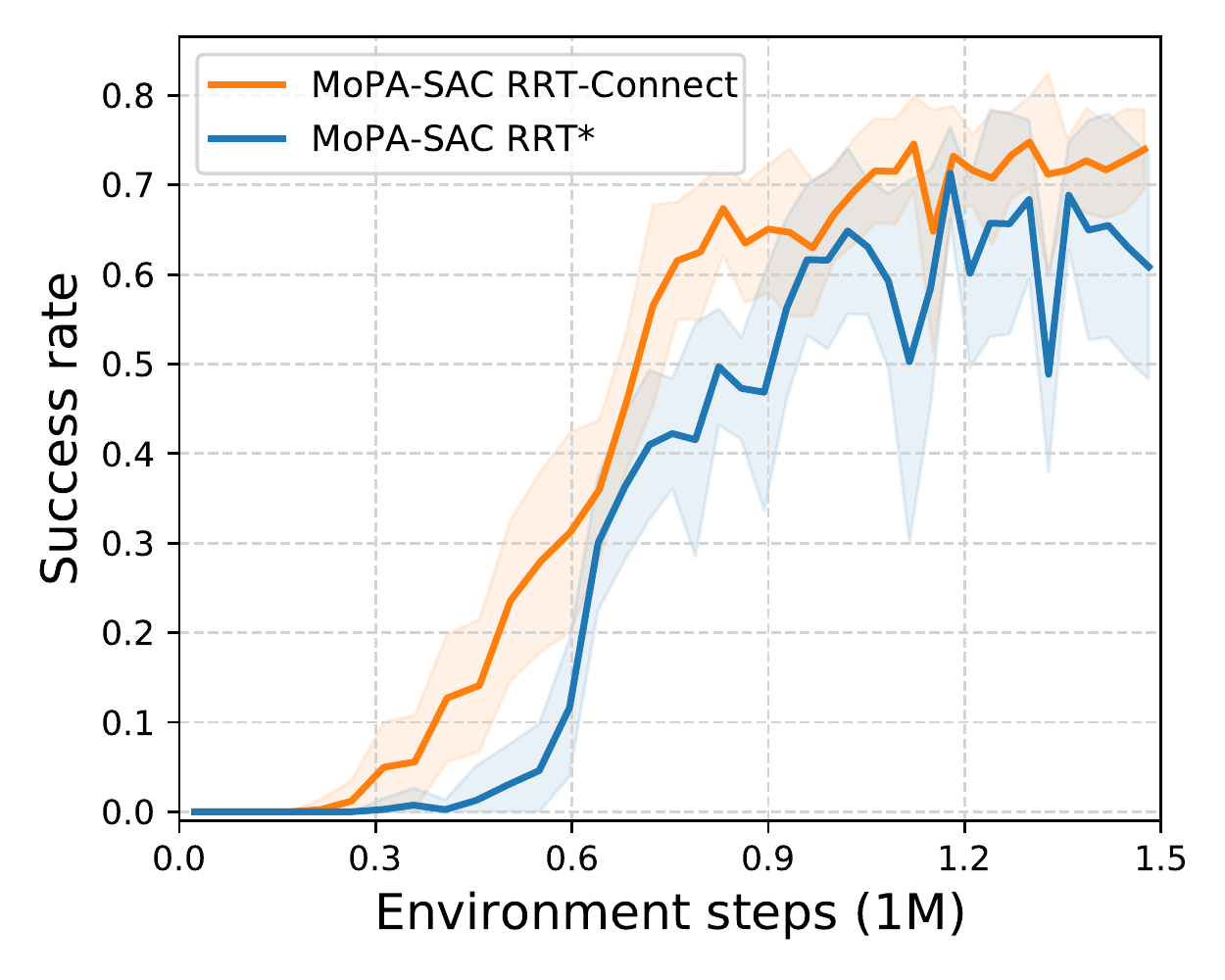}
      \caption{Different motion planners}
      \label{fig:ablation_MP}
  \end{subfigure}
  \quad
  \begin{subfigure}[t]{0.4\textwidth}
      \includegraphics[width=\textwidth]{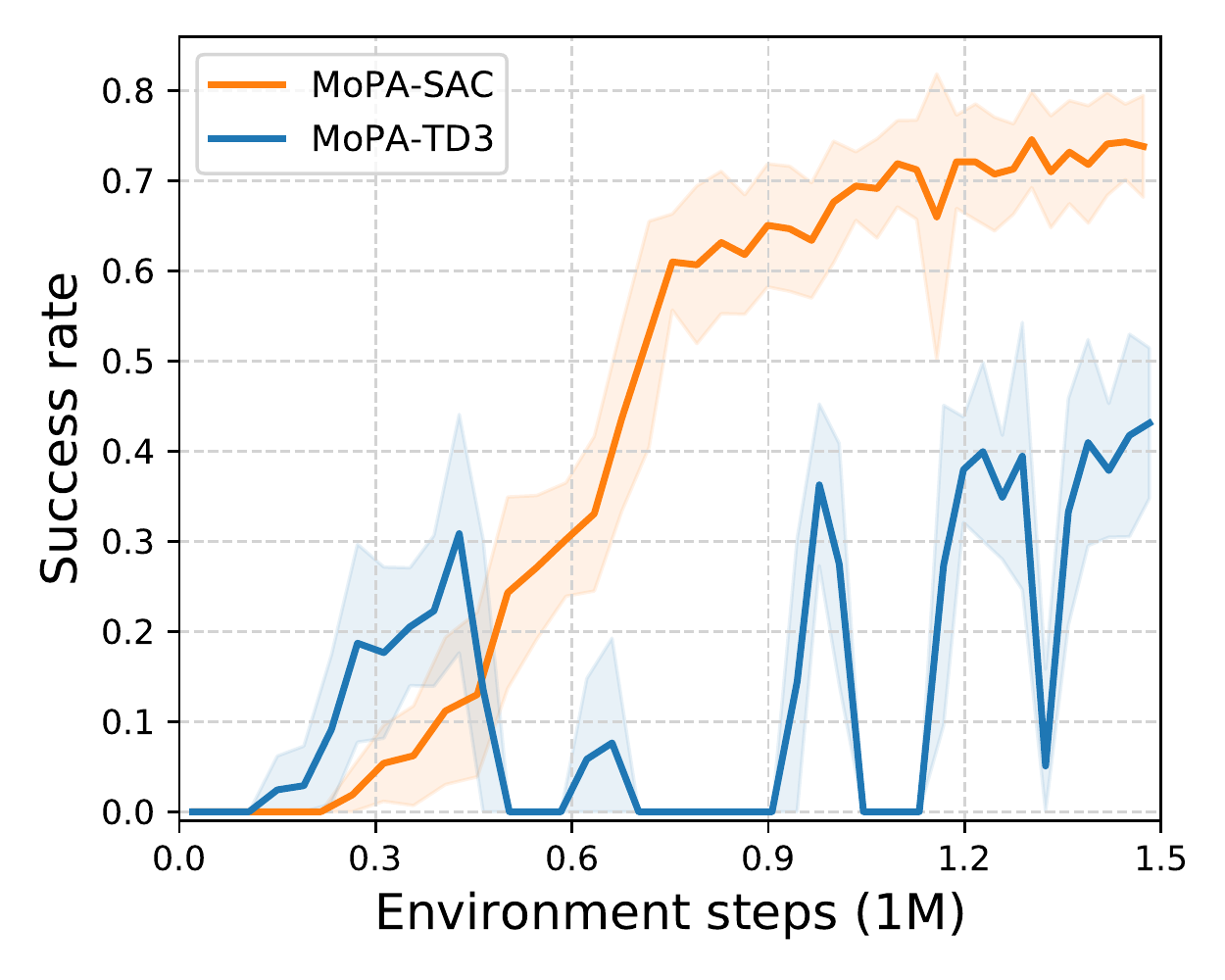}
      \caption{Different RL algorithms}
      \label{fig:ablation_RL}
  \end{subfigure}
  \caption{
    Learning curves of ablated models on \textit{Sawyer Assembly}. (a) Comparison of our model with different motion planner algorithms. (b) Comparison of our model with different RL algorithms.
  }
\end{figure}

To verify the compatibility of our method with different RL algorithms, we replaced SAC with TD3~\citep{td3} and compare the learning performance. As illustrated in \myfigref{fig:ablation_RL}, MoPA-TD3 shows unstable training, though the best performing seed can achieve around 1.0 success rate.

\section{Environment Details}
\label{supp:environment_details}

All of our environments are simulated in the MuJoCo physics engine~\citep{todorov2012mujoco}. The positions of the end-effector, object, and goal are defined as $p_\text{eef}$, $p_\text{obj}$, and $p_\text{goal}$, respectively. $T$ is the maximum episode horizon.

\begin{table}[ht]
\centering
\caption{Environment specific parameters for MoPA-SAC}
\vspace{0.5em}
\begin{tabular}{|c|ccccccc|} 
\toprule
Environment     & Action dimension & Reward scale & $\Delta q_\text{step}$ & $\Delta q_\text{MP}$ & $\omega$ & $M$ & $T$ \\ 
\midrule
2D Push        & 4 & 0.2 & 0.1 & 1.0 & 0.7 & 30 & 400\\ 
Sawyer Push     & 7 & 1.0 & 0.05 & 0.5 & 0.7 & 15 & 250 \\ 
Sawyer Lift     & 8 & 0.5 & 0.05 & 0.5 & 0.5 & 15 & 250 \\ 
Sawyer Assembly & 7 & 1.0 & 0.05 & 0.5 & 0.7 & 15 & 250\\ 
\bottomrule
\end{tabular}
\end{table}

\subsection{2D Push}

A 2D-reacher agent with 4 joints needs to first reach an object while avoiding obstacles and then push the object to the goal region. 

\textbf{Success criteria:} $||p_\text{goal}-p_\text{obj}||_{2} \leq 0.05$.

\textbf{Initialization:} The x and y position of goal and box are randomly sampled from $\mathcal{U}(-0.35, -0.24)$ and $\mathcal{U}(0.13, 0.2)$ respectively. Moreover, the random noise sampled from $\mathcal{U}(-0.02, 0.02)$ is added to the agent's initial pose.

\textbf{Observation:} The observation consists of $(\sin\theta, \cos\theta)$ for each joint angle $\theta$, angular joint velocity, the box position $p_\text{obj}=(x_\text{obj}, y_\text{obj})$, the box velocity, the goal position $p_\text{goal}$, and end-effector position $p_\text{eef}=(x_\text{eef}, y_\text{eef})$.

\textbf{Rewards:} Instead of defining a dense reward over all states which can cause sub-optimal solutions, we define the reward function such that the agent receives a signal only when the end-effector is close to the object (\ie $||p_\text{eef}-p_\text{obj}||_{2} \leq 0.1$). The reward function   consists of rewards for reaching the box and pushing the box to the goal region.
\begin{equation}
\begin{split}
    R_\text{push} = ~& 0.1\cdot \mathbbm{1}_{||p_\text{eef}-p_\text{obj}||_{2} \leq 0.1} (1-\tanh(5 \cdot ||p_\text{eef}-p_\text{obj}||_{2})) \\
    &+ 0.3\cdot \mathbbm{1}_{||p_\text{obj}- p_\text{goal}||_{2} \leq 0.1} (1-\tanh(5 \cdot ||p_\text{obj}-p_\text{goal}||_{2})) + 150 \cdot \mathbbm{1}_\text{success} 
\end{split}
\end{equation}

\subsection{Sawyer Push}

The \textit{Sawyer Push} task requires the agent to reach an object in a box and push the object toward a goal region.

\textbf{Success criteria:} $||p_\text{goal}-p_\text{obj}||_{2} \leq 0.05$.

\textbf{Initialization:} The random noise sampled from $\mathcal{N}(0, 0.02)$ is added to the goal position and the initial pose of the Sawyer arm.

\textbf{Observation:} The observation consists of each joint state $(\sin\theta, \cos\theta)$, angular joint velocity, the goal position $p_\text{goal}$, the object position and quaternion, end-effector coordinates $p_\text{eef}$, the distance between the end-effector and object, and the distance between the object and target.

\textbf{Rewards:}
\begin{equation}
\begin{split}
    R_\text{push} = ~&0.1\cdot \mathbbm{1}_{||p_\text{eef}-p_\text{obj}||_{2} \leq 0.1} (1-\tanh(5 \cdot ||p_\text{eef}-p_\text{obj}||_{2})) \\
    &+0.3\cdot \mathbbm{1}_{||p_\text{obj}-p_\text{goal}||_{2} \leq 0.1} (1-\tanh(5 \cdot ||p_\text{obj}-p_\text{goal}||_{2})) + 150 \cdot \mathbbm{1}_\text{success} 
\end{split}
\end{equation}

\subsection{Sawyer Lift}

In \textit{Sawyer Lift}, the agent has to pick up an object inside a box. To lift the object, the Sawyer arm first needs to get into the box, grasp the object, and lift the object above the box. 

\textbf{Success criteria:} The goal criteria is to lift the object above the box height.

\textbf{Initialization:} Random noise sampled from $\mathcal{N}(0, 0.02)$ is added to the initial position of a sawyer arm. The target position is always above the height of the box.

\textbf{Observation:} The observation consists of each joint state $(\sin\theta, \cos\theta)$, angular joint velocity, the goal position, the object position and quaternion, end-effector coordinates, the distance between the end-effector and object.

\textbf{Rewards:} 
This task can be decomposed into three stages; reach, grasp, and lift. For each of the stages, we define the reward function, and the agent receives the maximum reward over three values. The success of grasp is detected when both of the two fingers touch the object.

\begin{equation}
\begin{split}
    R_\text{lift} = & \max\big(\underbrace{0.1 \cdot (1-\tanh(10 \cdot ||p_\text{eef}-p_\text{obj}||_{2}))}_{\text{reach}}, \underbrace{0.35\cdot\mathbbm{1}_\text{grasp}}_{\text{grasp}},\\ & \underbrace{0.35\cdot \mathbbm{1}_\text{grasp}+0.15\cdot(1-\tanh(15\cdot \max(p^{z}_\text{goal}-p^{z}_\text{obj}, 0)))}_{\text{lift}}\big) + 150 \cdot \mathbbm{1}_\text{success} 
\end{split}
\end{equation}

\subsection{Sawyer Assembly}

The \textit{Sawyer Assembly} task is to assemble the last table leg to the table top where other three legs are already assembled.
The Sawyer arm needs to avoid the other table legs while moving the leg in its gripper to the hole since collision with other table legs can move the table. Note that the table leg that the agent manipulates is attached to the gripper; therefore, it does not need to learn how to grasp the leg.

\textbf{Success criteria:} The task is considered successful when the table leg is inserted into the hole. The goal position is at the bottom of the hole, and its success criteria is represented by $||p_\text{goal}-p_\text{leg-head}||_{2} \leq 0.05$, where $p_\text{leg-head}$ is position of head of the table leg.

\textbf{Initialization:} Random noise sampled from $\mathcal{N}(0, 0.02)$ is added to the initial position of the Sawyer arm. The pose of the table top is fixed.

\textbf{Observation:} The observation consists of each joint state $(\sin\theta, \cos\theta)$, angular joint velocity, the hole position $p_\text{goal}$, positions of two ends of the leg in hand $p_\text{leg-head}, p_\text{leg-tail}$, and quaternion of the leg.

\textbf{Rewards:}
\begin{equation}
    R_\text{assembly} = 0.4\cdot \mathbbm{1}_{||p_\text{leg-head}-p_\text{goal}||_{2} \leq 0.3} (1-\tanh(15 \cdot ||p_\text{leg-head}-p_\text{goal}||_{2})) + 150 \cdot \mathbbm{1}_\text{success} 
\end{equation}

\section{Training Details}
\label{supp:training_details}

For reward scale in our baseline, we use $10$ for all environments. In our method, each reward can be much larger than the one in baseline because it uses a cumulative reward along a motion plan trajectory when the motion planner is called. Therefore, larger reward scale in our method degrades the performance, and using small reward scale $0.1 \sim 0.5$ enables the agent to solve tasks. Moreover, $\alpha$ in SAC, which is a coefficient of entropy, is automatically tuned. To train a policy over discrete actions with SAC, we use Gumbel-Softmax distribution~\citep{jang2016categorical} for categorical reparameterization with temperature of $1.0$.

\begin{table}[h]
\centering
\caption{SAC hyperparameter}
\vspace{0.5em}
\begin{tabular}{|c|c|} 
 \toprule
 Parameter & Value  \\ 
 \midrule
 Optimizer & Adam  \\ 
 Learning rate & 3e-4\\
 Discount factor ($\gamma$) & 0.99 \\
 Replay buffer size & $10^6$ \\
 Number of hidden layers for all networks & 2 \\
 Number of hidden units for all networks & 256 \\
 Minibatch size & 256 \\
 Nonlinearity & ReLU \\
 Target smoothing coefficient ($\tau$) & 0.005 \\
 Target update interval & 1 \\
 Network update per environment step & 1 \\
 Target entropy & $-\text{dim}(\mathcal{A}$) \\
 \bottomrule
\end{tabular}
\end{table}

\subsection{Wall-clock Time}

The wall-clock time of our method depends on various factors, such as the computation time of an MP path and the number of policy updates. As Table \ref{tabel:wall-clock} shows, MoPA-RL learns quicker in wall-clock time compared to SAC for 1.5M environment steps. This is because SAC updates the policy once for every taken action, and our method requires fewer policy actions for completing an episode. As a result, our method performs fewer costly policy updates. Moreover, while a single call to the motion planner can be computationally expensive (~0.3 seconds in our case), we need to invoke it less frequently since it produces a multi-step plan (40 steps on average in our experiments). We further increased the efficiency of our method by introducing a simplified interpolation planner.

\begin{table}[h]
\centering
\caption{Comparison of the wall-clock training time in hours}
\vspace{0.5em}
\begin{tabular}{|c|c|c|c|} 
 \toprule
  & Sawyer Push & Sawyer Lift & Sawyer Assembly  \\ 
 \midrule
 MoPA-SAC & 15 & 17  & 14 \\ 
 SAC & 24 & 24 & 24\\
 \bottomrule
\end{tabular}
\label{tabel:wall-clock}
\end{table}

\end{document}